\documentclass{bmvc2k}
\usepackage{enumitem}
\usepackage{times}
\usepackage{epsfig}
\usepackage{graphicx}
\usepackage{amsmath}
\usepackage{amssymb}
\usepackage{longtable}
\usepackage{capt-of}
\usepackage[linesnumbered,ruled,vlined,boxed]{algorithm2e}

\usepackage{multicol}
\usepackage{multirow}
\usepackage{pifont}

\usepackage{bm}

\newcommand{\cmark}{\ding{51}}
\newcommand{\xmark}{\ding{55}}

\newcommand{\Te}{\emph{Teacher}}

\newcommand{\St}{\emph{Student}}

\SetCommentSty{mycommfont}

\title{\textit{Beyond Classification:} Knowledge Distillation using Multi-Object Impressions}
\addauthor{Gaurav Kumar Nayak*, Monish Keswani*}{{gauravnayak,monishkumar}@iisc.ac.in}{1}
\addauthor{Sharan Seshadri, Anirban Chakraborty}{sharan2510@gmail.com, anirban@iisc.ac.in}{1}

\addinstitution{
 Indian Institute of Science\\
 Bangalore, India\\
}
\runninghead{Nayak et al.}{Knowledge Distillation with MOIs}

\def\eg{\emph{e.g}\bmvaOneDot}

\def\etal{\emph{et al}\bmvaOneDot}
\begin{document}

\maketitle
\vspace{-0.2in}
\begin{abstract}

   Knowledge Distillation (KD) utilizes training data as a transfer set to transfer knowledge from a complex network (Teacher) to a smaller network (Student). Several works have recently identified many scenarios where the training data may not be available due to data privacy or sensitivity concerns and have proposed solutions under this restrictive constraint for the classification task. Unlike existing works, we, for the first time, solve a much more challenging problem, i.e., ``KD for object detection with zero knowledge about the training data and its statistics''. 
   Our proposed approach prepares pseudo-targets and synthesizes corresponding samples (termed as ``\textit{Multi-Object Impressions}''), using only the pretrained Faster RCNN Teacher network. 
   We use this pseudo-dataset as a transfer set to conduct zero-shot KD for object detection. We demonstrate the efficacy of our proposed method 
   through several ablations and extensive experiments on benchmark datasets like KITTI, Pascal and COCO. Our approach with no training samples, achieves a respectable mAP of $64.2$\% and $55.5\%$ on the student with same and half capacity while performing distillation from a Resnet-$18$ Teacher of $73.3\%$ mAP on KITTI.
\end{abstract}

\vspace{-0.3 in}
\section{Introduction}
\label{sec:intro}
\vspace{-0.1in}
Object detection has been an important and active area of research in the computer vision community. It deals with assigning a class label and a bounding box to each object in a given image. It is widely used across several applications. For example, in autonomous cars~\cite{chen2016monocular, wang2019pseudo} where correctly localizing the traffic signals, signboards, pedestrians, etc., is crucial to avoid accidents. They have also been used to analyze aerial images~\cite{yang2019clustered, vsevo2016convolutional} and perform multi-object tracking~\cite{Chen2018RealTimeMP, Wojke2017SimpleOA}. As detection is a crucial component in several vision-based applications, most of the research works primarily focus on making the object detection models as accurate as possible by leveraging deep networks and large amounts of training data. Such models are not suitable for deployment on portable devices that have limited memory and computational power. 
Therefore, there is an essential requirement to make such models compact and fast while retaining high accuracy.

Several compression techniques exist in the literature for obtaining a lightweight model from a complex deep model like pruning~\cite{han2015learning, srinivas2015data, Luo_2017_ICCV}, quantization~\cite{krishnamoorthi2018quantizing} and  low-rank factorization~\cite{sainath2013low}.  Their limitations are: i) architecture-dependence, ii) heuristics-based, iii) drop in accuracy. On the other hand, Knowledge Distillation~\cite{hinton2015distilling} (KD) transfers the knowledge from a trained large network (\Te{}) to a smaller network (\St{}) by matching the temperature raised soft labels along with cross-entropy loss on the ground truth. This ``dark knowledge'' provided by the teacher helps the student models to generalize well, without much drop in accuracy. Also, it is architecture-independent and does not require any heuristics. So, we restrict ourselves to KD as a means to learn compact models. 

In general, KD uses training dataset on which \Te{} was trained, as a transfer set for knowledge transfer from \Te{} to \St{}. 
However, we may only have access to the trained models and not their training data as several companies may have proprietary rights over them (\eg pretrained models~\cite{jeon2020compensating} on Google’s JFT-300M~\cite{kolesnikov2020big} proprietary dataset).
Also, if the data is sensitive (\eg medical and biometric data), it may not be shared due to privacy concerns. 
Recently several works have identified such issues~\cite{nayak2019zero, micaelli2019zero, chen2019data} for classification setting. 
The solutions proposed for classification problems either synthesize transfer set directly using the trained Teacher model~\cite{nayak2019zero, dreaming-to-distill-cvpr-2020} or learn the target data distribution through generative models~\cite{micaelli2019zero, chen2019data}. However, both of these approaches cannot be readily applied for object detection which is a more challenging and difficult problem. 
Object detection has dual priorities of object classification along with localization, and at the same time each image may contain variable number of objects. Moreover, each object can be of different spatial scales and aspect ratios, which can be present at various locations in an image. 

\begin{figure*}[htp]
\centering
\centerline{\includegraphics[width=\textwidth, height=0.38\textwidth]{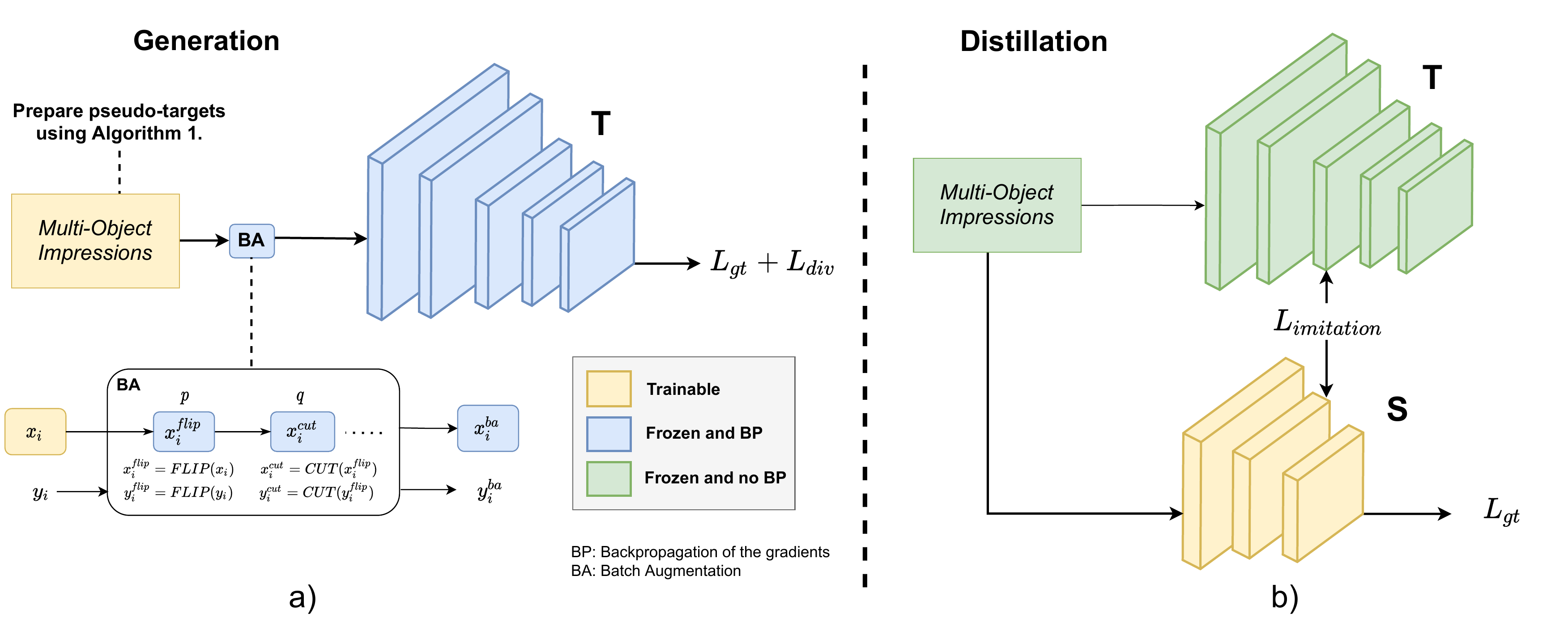}}
\scriptsize
\vspace{-0.1 in}
\caption{\small{Our proposed approach for zero-shot kd on object detection: a. Generation (left) where pseudo-dataset is synthesized using Algo. 2.; b. Distillation (right) is done using our pseudo-dataset as a transfer set. 
Note: we use our prepared pseudo-targets as ground truth to calculate the loss values.}}
\label{fig:arch}
\end{figure*}

Several works\cite{chen2017learning, wang2019distilling, tang2019learning} in object detection, assume the availability of the training data on which \Te{} is trained and uses them to distill the knowledge to \St{}. However, to perform KD for object detection in the absence of training data is a non-trivial and challenging problem. In our current work, we addressed this problem by synthesizing pseudo-dataset using only the trained Faster RCNN based \Te{} detection network and use them as a transfer set to conduct knowledge transfer in the data-free setting. Our proposed approach is broadly divided into two phases: Generation and Distillation as shown in Fig.~\ref{fig:arch}. 

In the generation phase, we first prepare pseudo-targets using proposed Algo.~\ref{algo_target} in absence of the training data and its statistics. 
Every object has three attributes associated with it : size, location and class label. We use the anchor scales and ratios obtained from \Te{} model to estimate the range of the object sizes. Based on the number of objects 
required to be placed in a given image dimension 
with a given IOU overlap constraint, we restrict the minimum and maximum possible areas for each object. We use Power Law distribution to sample the object area from this range. As the class distribution of objects is not known, we uniformly sample the class labels for each object.

The second step in the generation phase is to synthesize inputs corresponding to the prepared pseudo-targets (details in proposed Algo.~\ref{algo_MOIs}). We use random texture images as background initialization and optimize them by backpropagating the gradients of Faster RCNN loss ($L_{gt}$) and our proposed diversity loss ($L_{div}$) via the frozen \Te{} model. 
Our generated samples named as \textit{Multi-Object Impressions} (MOIs) are impressions of multiple objects of same/different classes at different locations with different scales
. We further make our generated MOIs invariant to augmentation operations like flipping, cutout via differential batch augmentation, and the proposed diversity loss further improves the intraclass variation on objects (shown in Fig.~\ref{fig:diversity}). 
In the distillation phase, we use our pseudo-dataset as transfer set and perform zero-shot KD. Our generated data is even suitable to be used beyond transfer set as we obtain reasonable 
mAP even while training the network from scratch with our data.

We thus, summarize our contributions as follows:
\begin{itemize}[noitemsep,topsep=0pt]
\itemsep0em
\item We are the first to attempt knowledge distillation on object detection, assuming zero knowledge about the training data and their statistics.

\item We propose an algorithm to synthesize pseudo-dataset comprising of target labels and corresponding input data, i.e. Multi-Object Impressions using a trained deep Faster RCNN detection model. Our pseudo-dataset is robust to augmentation operations.
\item We show the utility of such pseudo-dataset as a transfer set for distillation and are even suitable to train a detection network without any \Te{} assistance. Moreover, they can be used as augmentation in the presence of proxy data or few training samples.
\item We propose a diversity loss that enhances the intraclass variation on the foreground objects of MOIs
, leading to an improved distillation performance.


\item The effectiveness of our proposed approach is demonstrated across several architectures and benchmark datasets like KITTI, Pascal and COCO. 
\end{itemize}
\vspace{-0.33in}
\section{Related Works}
\label{sec:related}
\vspace{-0.1in}

\hspace{0.18in}\textbf{KD for classification:} 
The task of KD for the image classification, involves training a \St{} network on a temperature-raised softmax output of the teacher network. In the past years, the trend in this domain has been towards using as less data as possible. Kimura \etal~\cite{kimura2018few} used few training samples and generated pseudo-examples for KD 
, whereas Lopes \etal~\cite{dfkd-nips-lld-17} stored feature activations of all layers and used them in the form of meta-data. 
Nayak \etal~\cite{nayak2019zero} proposed a data-free approach to KD, where the training set was composed of data-impressions obtained via modeling the softmax output space through Dirichlet distribution. Micaelli \etal~\cite{micaelli2019zero} proposed an adversarial method that trains a generator iteratively to craft images that cause the student to poorly match the teacher and subsequently used them to perform distillation. Chen \etal~\cite{chen2019data} used an adversarial generator to synthesize images such that the teacher network gives high feature response and classifies them with high probability. Few works~\cite{yin2020dreaming, haroush2020knowledge} also uses batchnorm statistics of the trained classifier to regularize the feature distribution. Li \etal~\cite{li2021mixmix} uses features from several pretrained models and data mixup strategy to avoid model bias while matching such feature statistics. 
Thus, the problem of implementing a zero-shot/data-free approach to KD for image classification has produced a diverse set of solutions, each with a distinctive approach.

\textbf{KD for object detection:} 
Chen \etal~\cite{chen2017learning} proposed a distillation framework for object detection that used the Faster-RCNN model, which acts as a baseline model in a data-driven distillation framework. This approach used hint learning~\cite{romero2014fitnets} to improve the feature representation capability of the student, after which the classification and regression outputs of both the proposal network and the region CNN were distilled to the student. Wang \etal~\cite{wang2019distilling} avoid full imitation to overcome the noise introduced from various background instances and use fine-grained feature imitation exploiting important information in the near object anchor locations, which helps the Teacher to generalize better. All of these approaches in detection use large training datasets, often with several classes to perform distillation.

One way to handle the absence of training data is to generate synthetic data and use them for the downstream task. Few works such as \cite{georgakis2017synthesizing, tobin2017domain, hinterstoisser2018pre} generate synthetic data 
using external tools like CAD or depend on the availability of cropped images of objects. Therefore, there is an implicit assumption of familiarity with the target categories. In absence of such priors, synthetic samples are generated from a pretrained Yolo network in \cite{chawla2021data} but their method depends on training data statistics in the form of batch norm. Also, their focus is on one-stage detectors with the goal of only knowledge transfer. 

We, in this paper, focus on two stage-detectors like Faster RCNN with an objective of model compression along with knowledge transfer where zero knowledge about training samples and training data statistics are assumed. 
\vspace{-0.2in}
\section{Proposed Approach}
\label{sec:proposed}
\vspace{-0.1in}
Our approach for zero-shot KD for object detection is broadly divided into two phases (also shown in Figure~\ref{fig:arch}):
\begin{itemize}[noitemsep,topsep=0pt]
\itemsep0em
\item \textbf{Generation}: In order to synthesize samples in absence of training data and their statistics, we first need to prepare `pseudo-targets’ (in Sec.~\ref{subsec:labels}). We generate Multi-Object Impressions (MOIs) corresponding to them, using the pretrained weights of \Te{} network via backpropagation (in Sec.~\ref{subsec:generation}).
\item \textbf{Distillation}: The generated dataset ($\hat{D}$) can then be used as a transfer set to distill knowledge into the \St{} network (in Sec.~\ref{subsec:distillation}).
\end{itemize}

We define our pseudo-dataset as $\hat{D}=\{ (x_{i}, y_{i})\}_{i=1}^{K}$, where $K$ is the number of samples to be generated, $x_{i}$ is the $i^{th}$ input with corresponding pseudo-target $y_{i}$. The pseudo-target is defined as $y_{i} = \{(c_{io}, bbox_{io})\}_{o=1}^{N_{i}}$, where $N_{i}$ is the number of objects in input $x_{i}$, $c_{io}$ denotes target class and $bbox_{io}$ denotes target bounding box coordinates, for the $o^{th}$ object in $x_{i}$.
\begin{algorithm}[htp]
{
\footnotesize{
\caption{\footnotesize{Preparation of $B$ pseudo-targets for pseudo-dataset $\hat{D}$ using \Te{} $T$}}
\label{algo_target}
\SetAlgoLined
\SetKwInOut{Input}{Input}  
\Input{$W_{i}, H_{i}$: width and height of $i^{th}$ input, \ \ $IoU_{thresh}$: max IoU threshold, \ \  $C$: $\#$ foreground classes \\
$t_{1}, t_{2}$: control $A^{max}$ and $A^{min}$ respectively,
\ $R_{min}, R_{max}$: min and max aspect ratio,\\ 
}
\SetKwInOut{Output}{Output}  
\Output{
$B$ Pseudo-Targets: $\{y_{i}\}_{i=1}^{B}$}
$A^{min}, \ A^{max}$ $\leftarrow$ Minimum and Maximum object area based on range of anchor scales and $t_{1}, t_{2}$ \label{a_minimax} \\
Choose $M$ such that $M\le(W_{i}\cdot H_{i})/A^{min}, \forall i\in [1..B]$ \label{alg:M} 
\\
\For{$i=1:B$}{
$y_{i} \leftarrow \varnothing$; \ $cur\_iter \leftarrow 0$; \ $max\_iters \leftarrow 50$ \\
$N_{i} \sim \mathcal{U}\{1, M\}$; \  $a_{i} \leftarrow M / N_{i}$\\
$A_{i}^{max} \leftarrow \min(A^{max}, (W_{i}\cdot H_{i}) / N_{i})$\\ \label{a_i_max}
\For{$o=1:N_{i}$}{
$c_{io} \sim \mathcal{U}\{1, C\}; \ r_{io} \sim \mathcal{U}(R_{min},R_{max})$\\
\While{$cur\_iter \leq max\_iters$}{
$x_{io}\sim P(a_{i})$, where $P(x; a) = ax^{a-1}, 0 \le x \le 1, a>0$\label{size_1} \\ 
$A_{io} \leftarrow (A_{i}^{max} - A^{min})\cdot x_{io}+A^{min}$\\
Obtain width $w_{io}$ and height $h_{io}$ using area $A_{io}$ and aspect ratio $r_{io}$ \label{size_2}\\
$[\ ctr_{io}^{x}, \ ctr_{io}^{y}] \sim  [\ \mathcal{U}\{w_{io}/2, W_{i} - w_{io}/2+1\}, \ \mathcal{U}\{h_{io}/2, H_{i} - h_{io}/2+1\}$] \label{size_3} \\ 
$bbox_{io} \leftarrow [ctr_{io}^{x}, ctr_{io}^y, w_{io}, h_{io}]$\\ 
\eIf{ (IoU( $bbox_{io}$, $bbox$) $< IoU_{thresh}$ $\forall bbox \in y_{i}$) or ($cur\_iter == max\_iters$) }{
$y_{i} \leftarrow y_{i} \cup \{(bbox_{io}, c_{io})\}$; \ break\\
}
{
$cur\_iter \leftarrow cur\_iter + 1 $
} \label{size_4}
}
}
}
}
}
\end{algorithm}
\vspace{-0.15in}
\subsection{Preparation of Pseudo-Targets}
\label{subsec:labels}
In detection, multiple objects can be present in an image. These objects can be of variable sizes, at different locations and belong to same or different classes which makes the problem complex. We aim to prepare pseudo-targets that are close to the annotations of training data in terms of sizes, locations and class labels of objects. 

The \textbf{size} of an object is defined in terms of scales and aspect ratios~\cite{ren2015faster}. We leverage on the anchor ratios and scales obtained from the pretrained detection model to get some insight regarding the range of object sizes in the training dataset. Please note that it only provides a clue about the estimated range of the object sizes in the training dataset but in no way reveals their actual sizes or the class labels. 

The \textbf{location} of an object is dependent on the other object's locations. We need to take care of their overlap so that one object does not get placed over other objects or have very high overlap. It is desirable to have the overlap within some predefined threshold. We use IoU measure to enforce such a constraint. 

The \textbf{class label} distribution of the training data is unknown and no prior information or metadata is available, so we use uniform sampling to assign a class label to an object. Algorithm~\ref{algo_target} contains the detailed steps to obtain pseudo-targets for pseudo-dataset ($\hat{D}$).

We define the minimum and maximum object area as $A^{min}$ and $A^{max}$ based on the anchor scales defined by the teacher model (Line \ref{a_minimax}). The maximum number of objects that are allowed in any sample is denoted by $M$. For an $i^{th}$ image ($x_{i}$), the important factors on which its pseudo-target depends are: the number of objects, class label, size (aspect ratio and area) and location of each object in the image. We now discuss in detail how our proposed algorithm handles each of these factors.

The number of objects is sampled uniformly i.e. $N_{i} \sim \mathcal{U}\{1, M\}$. The class label $c_{io}$ is assigned to the $o^{th}$ object using uniform distribution i.e. $c_{io} \sim \mathcal{U}\{1, C\}$ where $C$ is the number of foreground classes. The aspect ratio of each object ($r_{io}$) is uniformly sampled from the range of anchor ratios defined by the teacher model. Based on the dimensions of $x_{i}$ and value of $N_{i}$, we constrain the maximum possible object area and denote it by $A^{max}_{i}$(Line~\ref{a_i_max}). We utilize Power Law distribution $P$ to sample from the range [$A^{min}$, $A_{i}^{max}$], denoted by $A_{io}$. We define the parameter $a_{i}$ as a function of $N_{i}$ (Refer to supplementary for details). 
Using $A_{io}$ and $r_{io}$, width ($w_{io}$) and height ($h_{io}$) of the target bounding box are obtained for the $o^{th}$ object in $x_{i}$ (Lines~\ref{size_1}-~\ref{size_2}). Next (Lines~\ref{size_3}-~\ref{size_4}), we place the $o^{th}$ object in $x_{i}$ such that object overlap between each pair is less than the IoU threshold. The threshold condition often gets satisfied within a few iterations. Otherwise we save the pseudo-targets after maximum iterations.
\vspace{-0.2 in}
\subsection{Crafting MOIs for Pseudo-Targets}
\label{subsec:generation}
Let the pretrained Faster RCNN model be denoted by $T$ and its learnt model parameters by $\theta_{T}$ which is trained on a detection training dataset ($D$). In the absence of the dataset $D$, we aim to synthesize pseudo dataset $\hat{D}$. 
After obtaining the pseudo-target $y_{i}$ using proposed algorithm~\ref{algo_target}, we need to synthesize $x_{i}$ using pretrained weights of $T$ such that:
\begin{equation}
\label{eq1}
    T(x_{i}) = y_{i},  \forall i \in [1..K]
\end{equation}
In order to satisfy this condition, every data sample $x_i$ requires to have impressions of objects of variable sizes at different locations belonging to same or different classes. Therefore, we call such $x_{i}$'s as Multi-Object Impressions (MOIs). In other words, the $i^{th}$ sample of pseudo-dataset $\hat{D}$ corresponding to $i^{th}$ pseudo-target ($y_{i}$) is the $i^{th}$ MOI (denoted by $x_{i}$). 

The RCNN based detectors classify the diverse backgrounds in images into a single background class. To handle this background variability, we initialize our MOIs with samples from a texture dataset~\cite{cimpoi2014describing}. In comparison to random noise, 
feature maps extracted by the layers of base network of pretrained detector have higher activation values and acts as better initialization.
The texture images do not contain any objects. Also, the \Te{} network predicts background class on such images with high confidence. Thus, the initialization of each $x_{i}$’s with texture image act as background initialization for the MOIs. 


Data augmentation~\cite{zoph2019learning} is a commonly used technique to improve the performance of the training model. But we cannot directly apply data augmentation on MOIs during KD, since MOIs are the samples on which the \Te{} model is confident. We cannot predict the behaviour of the \Te{} on the augmented MOIs. The \Te{} may not predict the required pseudo-target on the augmented MOIs and can violate equation~\ref{eq1}. So, we make the MOIs robust to augmentation operations via differentiable batch augmentation. 

Batch augmentation~\cite{Hoffer2019AugmentYB} (BA) improves the generalization of the model and also leads to faster convergence. We use this approach\footnote{\label{batch_aug}performs augmentation on MOIs in a batch 
} to make our MOIs invariant to certain augmentations. 
We perform two differential batch augmentations, namely Flip and Cutout~\cite{devries2017improved}. In BA module in Fig.~\ref{fig:arch}, $x_{i}$ (i.e. $i^{th}$ MOI) is passed through differentiable augmentation operations with probability $p$ and $q$. We set $p=q=0.5$ and $x_{i}^{ba}$ is the output of BA module which is then fed to the \Te{} network.

After the initialization and augmentation, we optimize each $x_{i}$ keeping $\theta_{T}$ fixed with target labels $y_{i}, \forall i\in [1..K]$ using the loss $L$ as defined:
 \begin{equation}
     L_{MOI} = \lambda_{gt} L_{gt} + \lambda_{div} L_{div}
 \end{equation}
where $\lambda_{gt}$ and $\lambda_{div}$ are hyperparameters which are used to balance the losses. The description of the losses are mentioned below: \newline
\textbf{Detection Loss ($L_{gt}$)}: The usual classification and bounding boxes losses~\cite{ren2015faster} applied at RPN and RCNN layers in training the network $T$. In absence of original training dataset ($D$), we use our pseudo-dataset ($\hat{D}$). \newline
\textbf{Diversity Loss ($L_{div}$)}: To ensure intraclass variability across the foreground objects, we define diversity loss as:
\begin{equation}
    L_{div} = - \frac{1}{C} \sum_{c=1}^{C} \frac{1}{|N_{c}|} \sum_{(i, j)\in N_{c}} d(f_{i}, f_{j})
\end{equation}
where $C$ denotes the number of foreground classes, $N_{c}$ denotes the collection of pairs of foreground objects belonging to class $c$ in the current batch. $f_{i}$ and $f_{j}$ denote the pooled feature vectors of foreground objects $i$ and $j$ that also belongs to same class $c$. $d$ is distance metric (euclidean, cosine). We use cosine similarity for the experiments.
\begin{algorithm}[htp]
{\footnotesize
\caption{\footnotesize{Generation of pseudo-dataset ($\hat{D}$)}}
\label{algo_MOIs}
\SetAlgoLined
\SetKwInOut{Input}{Input}  
\Input{$T$: Pretrained Faster RCNN model, $K$: Number of samples,
$N$ : Number of iterations\\
}
\SetKwInOut{Output}{Output}  
\Output{$\hat{D}$ : pseudo-dataset}

$\hat{D} \leftarrow \varnothing$\\
Select batch size $b$, s.t. $K \bmod b = 0, b > 1$ \\
\For{$K/b$ \textit{batches}}{
Sample a minibatch of $b$ background images, $\{x_{1},x_{2}, \dots, x_{b}\}$ from DTD texture data~\cite{cimpoi2014describing}

Obtain a minibatch of $b$ pseudo-targets, $\{y_{1},y_{2}, \dots, y_{b}\}$ using Algo.~\ref{algo_target}\\

Associate the pseudo-targets with sampled images
$(x, y) = \{(x_{1}, y_{1}),(x_{2}, y_{2}), \dots, (x_{b}, y_{b})\}$\\
\For{$N$ iterations}{
Perform Batch augmentation 
$(x^{ba}, y^{ba}) \leftarrow \textit{BA} (x, y)$ \\
Update $x$ by descending its gradient
$\nabla_{x} L_{MOI}(T(x^{ba}), y^{ba})$
}
$\hat{D} \leftarrow \hat{D} \cup (x, y)$
}
}
\end{algorithm}
\newline\textbf{Training Setup}: MOIs initialized with texture images are trainable. The gradients are backpropagated with respect to each $x_{i}$’s through the frozen \Te{}. Each $x_{i}$ is optimized for $N$ iterations to minimize the loss between the \Te{}’s prediction and pseudo-targets $y_{i}$’s. The RPN loss from $L_{gt}$ forces the region proposals to be near to the target bbox’s in $y_{i}$’s. Further RCNN loss corrects the proposals and predicts each target class in $y_{i}$’s with high confidence. This would eventually lead to having foreground objects’ impressions such that the \Te{} network when fed with optimized MOIs, predicts the desired pseudo-targets. 
Diversity loss encourages objects of a target class to have diverse features which helps it to improve KD performance. 
Refer to Figure~\ref{fig:visualization} for visualization of synthesized MOIs and supplementary for more visualizations. Algo.~\ref{algo_MOIs} contains overall steps involved in the generation of $\hat{D}$.
\vspace{-0.15in}
\subsection{Distillation using pseudo dataset ($\hat{D}$)}
\label{subsec:distillation}
Let the student model be denoted by $S$ and its trainable parameters by $\theta_{S}$. After obtaining the pseudo-targets and corresponding MOIs (using Algorithm~\ref{algo_MOIs}), we use our synthesized pseudo dataset ($\hat{D}$) as transfer set to perform knowledge transfer from $T$ to $S$ using the detection loss ($L_{gt}$) and feature imitation loss ($L_{imitation}$) used in \cite{wang2019distilling}. We evaluate the generalization ability of the student $S$ trained with transfer set $\hat{D}$ through mean average precision (mAP) on actual test samples and compare its performance with data dependent approaches.
\vspace{-0.2in}

\section{Experiments}
\label{sec:kitti}
\vspace{-0.1in}
We first use Resnet-$18$~\cite{he2016deep} Faster R-CNN model as \Te{} network trained on KITTI \cite{Geiger2012CVPR} benchmark dataset. 
The performance of the trained \Te{} model is $73.3\%$ mAP. The models are evaluated based on the split followed by \cite{cai2016unified, mao2017can, wang2019distilling} using official evaluation tool. 

Several design choices are possible in crafting MOIs like maximum number of objects allowed per sample ($M$) and the size of pseudo-data ($K$). We discuss the effect of the aforementioned major factors on distillation performance in subsequent sections. 
We fix the size of the MOIs as $600 \times 600$ dimensions. We take $t_{1}$, $t_{2}$ and $IOU_{thresh}$ as $1.2$, $0.8$ and $0.1$ respectively. We use Adam optimizer and Pytorch framework for all the experiments. Refer supplementary for hyperparameter details used in experiments. 

\vspace{-0.2in}
\subsection{Maximum objects per sample}
\label{max_objects}
\begin{figure}[htp]
  \begin{minipage}{0.33\linewidth}
    \includegraphics[width=0.9\linewidth]{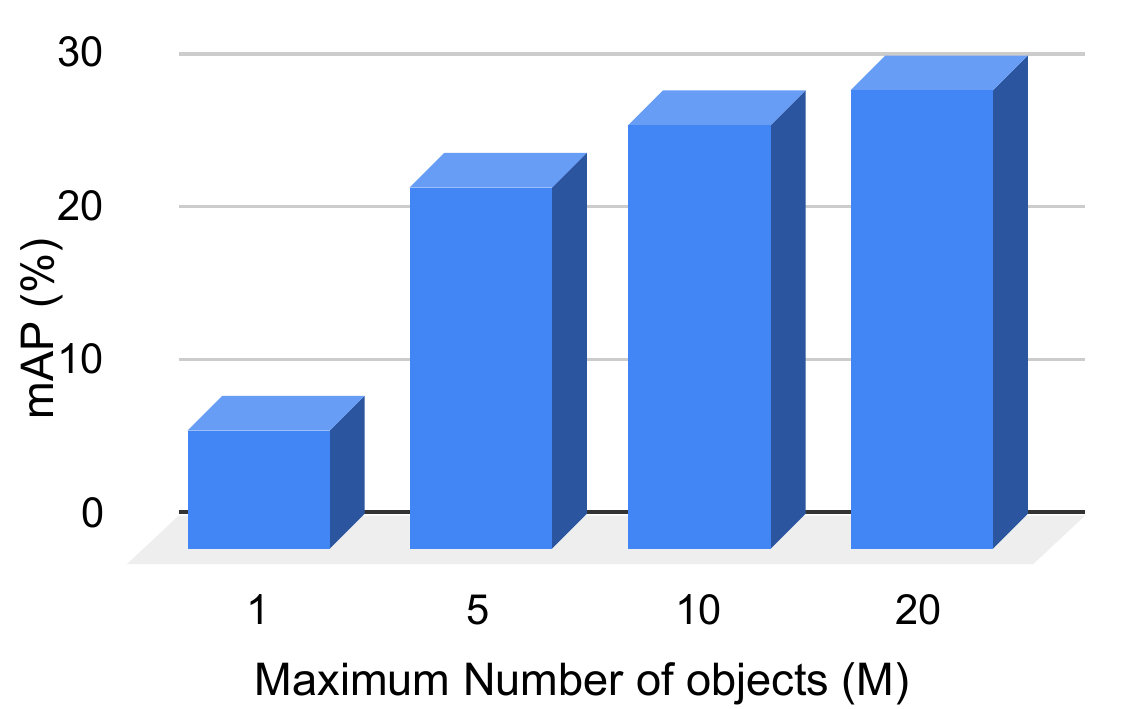}
    \vspace{-0.1in}
    \small
    \caption{\small{Comparison of distillation performance when maximum number of objects is varied.}}
    
    \label{fig:max_objects}
  \end{minipage}
  \hspace{.01\linewidth}
  \begin{minipage}{0.36\linewidth}
  
    \includegraphics[width=0.92\linewidth]{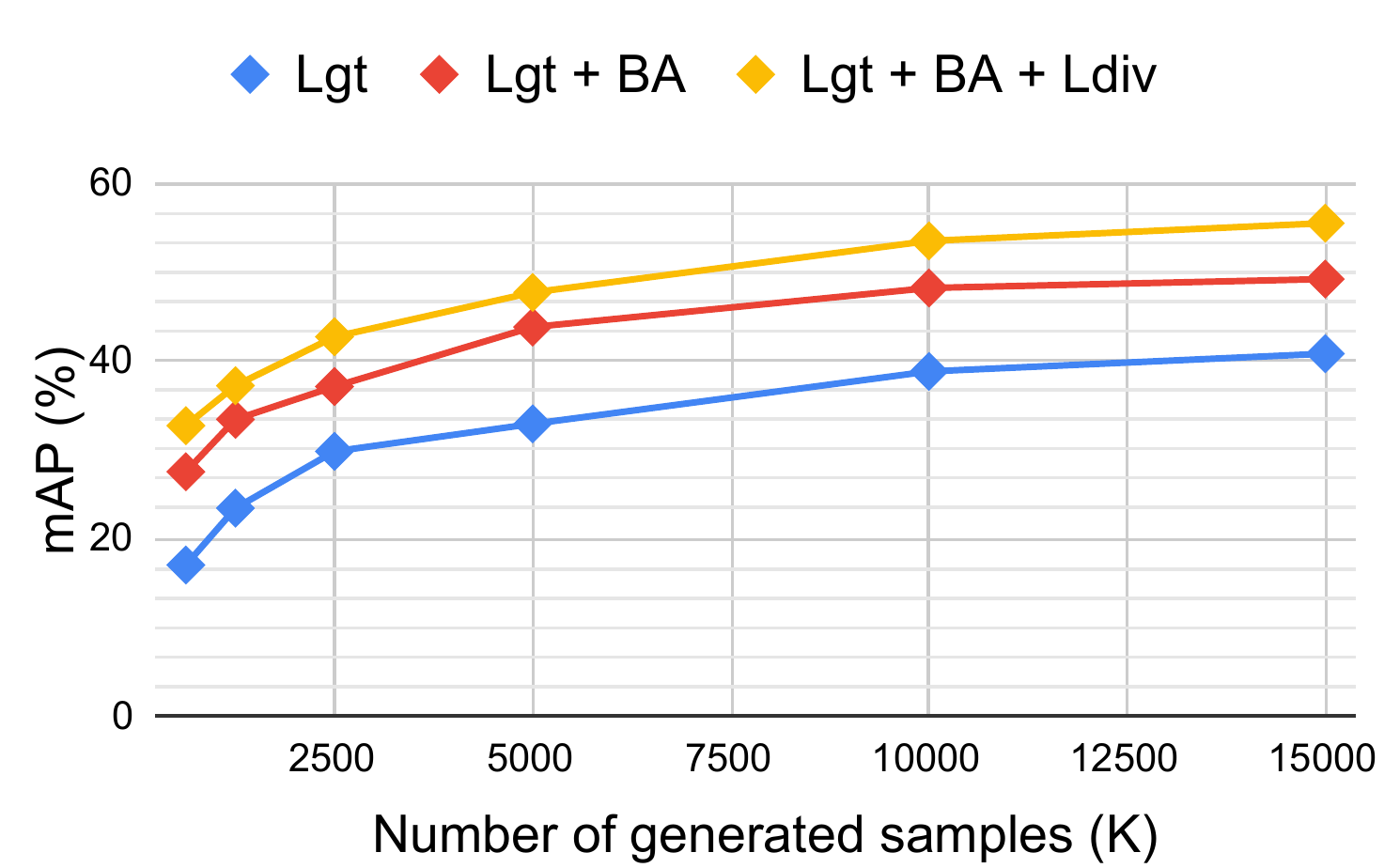}
    \vspace{-0.1in}
    \small
    \caption{\small{Detection accuracy (w/ KD) by varying number of samples, MOIs crafted via different losses.}}
    \label{fig:size}
  \end{minipage}
  \hspace{.01\linewidth}
  \begin{minipage}{0.26\linewidth}
  
    \includegraphics[width=0.98\linewidth, height=0.62\linewidth]{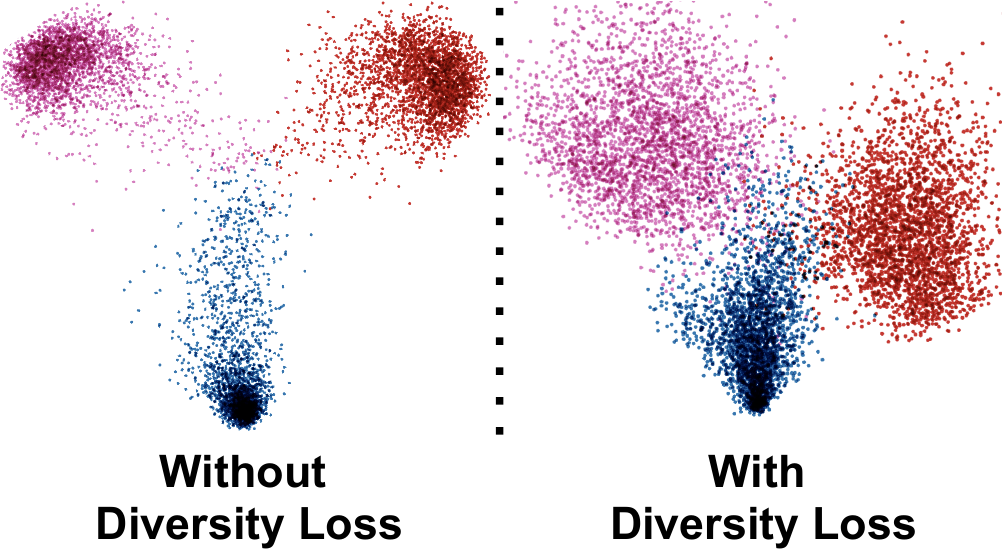}
    \vspace{-0.1in}
\small
\caption{\small{Visualization of pooled features of objects on the generated samples.
}}
\label{fig:diversity}
  \end{minipage}
  \vspace{-0.15in}
\end{figure}

We vary the maximum number of objects ($M$) possible in any sample of pseudo-dataset $\hat{D}$ and check its impact on distillation performance of Resnet-$18$-half student. It is evident from Figure~\ref{fig:max_objects} that mAP improves with an increase in the value of $M$. Large value of $M$ offers more variation in object sizes. However, we cannot have an arbitrary large $M$ due to the constraint specified in line no.~\ref{alg:M} of Algo.~\ref{algo_target}. Thus, we choose the value of $M$ as $20$ for subsequent experiments. 
These experiments are performed using transfer set size of $2500$ MOIs. We perform an ablation on the number of generated samples 
in the next section. 
%
\vspace{-0.15in}
\vspace{-0.4in}
\subsection{Size of pseudo-dataset}
\label{size}
We generate different sized pseudo-datasets by varying $K$ 
and analyze its effect while performing distillation on Resnet-$18$-half student. 
For a particular value of $K$, we generate MOIs using three different ways : a) using $L_{gt}$ loss, b) using $L_{gt}$ loss and batch augmentation (BA), c) using $L_{div}$ loss along with $L_{gt}$ loss and BA. 
The results are shown in Figure~\ref{fig:size}.

a) $\bm{L_{gt}}$ :  
The performance of the \St{} model with $L_{gt}$ (shown in blue curve) improves by increasing $K$ and achieves $40.8\%$ mAP with $15000$ pseudo-samples. 

b) $\bm{L_{gt}}$ \textbf{+ BA}: In order to make our MOIs robust to augmentations, we do simple differential batch augmentation like flipping and cutout during generation. This enforces the MOIs to be invariant to the augmentation operations. 
Robust MOIs indeed helps in improving the performance of the \St{} model as evident from Figure~\ref{fig:size} (shown in red curve). 
Though BA encourages robustness, but do not explicitly force the object impressions belonging to the same class to be diverse. 
So, in order to enforce high intraclass variation, we additionally add our proposed diversity loss ($L_{div}$).

c) $\bm{L_{gt}}$ \textbf{+ BA} \textbf{+} $\bm{L_{div}}$ : After adding $L_{div}$ loss on top of $L_{gt}$ and BA, we observe further gain in the performance of the \St{} model (shown in yellow curve in Figure~\ref{fig:size}).

\textbf{Proposed Diversity Loss} ($\bm{L_{div}}$): We visualize the pooled features of foreground object impressions for each class. From Fig.~\ref{fig:diversity} (left), we observe that without the diversity loss, the objects belonging to each class in the feature space lie close to their class means, and objects are well separated across classes. 
Through our proposed loss, we improve the intraclass variation as shown in Fig.~\ref{fig:diversity} (right). Also, sample density near the boundary region has increased which helps the \St{} to learn the decision boundary resulting in improved mAP.\\
\vspace{-0.34in}
\subsection{Results on KITTI} 
In Table~\ref{same_arch_half_channels}, we report our overall results 
while distilling from Resnet-$18$ ($73.3\%$ mAP) as \Te{} to Resnet-$18$ and Resnet-$18$-half as \St{}'s. We compare our data-free kd approach (no original training samples) with data kd approach (all original training samples) which serves as an upper bound. We also compare against the baseline that uses in-domain \cite{Cordts2016Cityscapes} and out-domain \cite{Ros_2016_CVPR} data and report their average performance. We get a significant improvement over the baseline. Our MOIs generated through a combination of $L_{gt}$ and $L_{div}$ losses using BA obtains distillation performance of $64.2\%$ and $55.5\%$ mAP on student with same and half capacity. Please note that even without KD, we observe decent mAP of $57.8\%$ by training Resnet-$18$ with our MOIs which shows its utility to be used beyond transfer set. 
\vspace{-0.1 in}
\subsection{Results on Pascal and COCO}
\label{sec:pascal}
Pascal VOC07 dataset~\cite{pascal-voc-2007} is another popular benchmark dataset for object detection tasks and is relatively more complex than KITTI dataset. It contains foreground objects that can belong to one of the $20$ classes. 
We perform distillation from two different \Te{} networks i.e. Resnet-$34$ and VGG-$16$ trained on Pascal. 
Similar to KITTI, we synthesize samples with $M$ as $20$, $K$ as $15000$ with batch augmentation, and optimize using $L_{gt}+L_{div}$ losses. The models are evaluated using Pascal VOC convention i.e. mAP at $0.5$ IoU. The results are shown in Table~\ref{tab:resnet}. 
Our proposed zero-shot kd method obtains a distillation performance of 
$55.2\%$ on VGG-$16$ and $58.6$\% mAP on Resnet-$34$. In absence of teacher assistance, we still get a decent performance of $49.3\%$ on VGG-$16$ and $50.9\%$ mAP on Resnet-$34$ when our pseudo-dataset is used to train the networks from scratch. 
\begin{table}[htp]
\begin{minipage}[htp]{0.5\textwidth}
  \scalebox{0.65}{
  \begin{tabular}{|c|c|c|c|c|c|}
\hline
\textbf{Setting} &
  \textbf{\begin{tabular}[c]{@{}c@{}}Training \\ Method\end{tabular}} &
  \textbf{\begin{tabular}[c]{@{}c@{}}Loss on\\ MOIs\end{tabular}} &
  \textbf{Teacher} &
  \textbf{Student} &
  \textbf{mAP} \\ \hline\hline
\multirow{3}{*}{\textbf{\begin{tabular}[c]{@{}c@{}}With \\ training \\ data\end{tabular}}} &
  w/o KD~\cite{ren2015faster} & \multirow{2}{*}{N/A} &
  \multirow{2}{*}{\begin{tabular}[c]{@{}c@{}}\\Resnet-18\end{tabular}} &
  -------- &
  73.3 \\ \cline{2-2} \cline{5-6} 
 &
  w/ KD~\cite{wang2019distilling} &
   &
   &
  \begin{tabular}[c]{@{}c@{}}Resnet-\\ 18-half\end{tabular} &
  65.8 \\ \hline\hline
\multirow{6}{*}{\textbf{\begin{tabular}[c]{@{}c@{}}Without \\ training \\ data\\ (Ours)\end{tabular}}} &
  Baseline &
  N/A &
  \multirow{4}{*}{\begin{tabular}[c]{@{}c@{}}Resnet-\\ 18\end{tabular}} &
  \multirow{4}{*}{\begin{tabular}[c]{@{}c@{}}Resnet-\\ 18-half\end{tabular}} &
  42.3 \\ \cline{2-3} \cline{6-6} 
 &
  w/ KD &
  Lgt &
   &
   &
  40.8 \\ \cline{2-3} \cline{6-6} 
 &
  w/ KD &
  Lgt+BA &
   &
   &
  49.2 \\ \cline{2-3} \cline{6-6} 
 &
  w/ KD &
  \multirow{3}{*}{\begin{tabular}[c]{@{}c@{}}Lgt\\ +BA\\ +Ldiv\end{tabular}} &
   &
   &
  55.5 \\ \cline{2-2} \cline{4-6} 
 &
  w/o KD &
   & --------
   &
  \multirow{2}{*}{\begin{tabular}[c]{@{}c@{}}Resnet-\\ 18\end{tabular}} &
  57.8 \\ \cline{2-2} \cline{4-4}\cline{6-6} 
 &
  w/ KD &
   & Resnet-18
   &
   &
  \textbf{64.2} \\ \hline
\end{tabular}}
\caption{\small{KD results using our proposed approach on KITTI dataset.}}
\label{same_arch_half_channels}
\end{minipage}\qquad
\begin{minipage}[htp]{0.45\textwidth}
  \scalebox{0.65}{
  \begin{tabular}{|c|c|c|c|c|}
\hline
\textbf{Setting} &
  \textbf{\begin{tabular}[c]{@{}c@{}}Training \\ Method\end{tabular}} &
  \textbf{Teacher} &
  \textbf{Student} &
  \textbf{mAP} \\ \hline\hline
\multirow{3}{*}{\textbf{\begin{tabular}[c]{@{}c@{}}With \\ training \\ data\end{tabular}}} &
  w/o KD~\cite{ren2015faster} &
  VGG-16 &
  -------- &
  70.4 \\ \cline{2-5} 
 & w/o KD~\cite{ren2015faster} & \multirow{2}{*}{Resnet-34} & --------                   & 70.1          \\ \cline{2-2} \cline{4-5} 
 & w/ KD~\cite{wang2019distilling}    &                            & Resnet-18                  & 67.8          \\ \hline\hline
\multirow{5}{*}{\textbf{\begin{tabular}[c]{@{}c@{}}Without \\ training \\ data\\(\textbf{Ours})\end{tabular}}} &
  w/o KD &
  -------- &
  \multirow{2}{*}{VGG-16} &
  49.3 \\ \cline{2-3} \cline{5-5} 
 & w/ KD  & VGG-16                     &                            & 55.2          \\ \cline{2-5} 
 & w/ KD  & Resnet-34                  & Resnet-18                  & 46.3          \\ \cline{2-5} 
 & w/o KD & --------                   & \multirow{2}{*}{Resnet-34} & 50.9          \\ \cline{2-3} \cline{5-5} 
 & w/ KD & Resnet-34                  &                            & \textbf{58.6} \\ \hline
\end{tabular}}
\caption{\small{Results of distillation 
on Pascal dataset using our proposed data-free approach.}}
\label{tab:resnet}
\end{minipage}
\end{table}
\vspace{-0.2in}
\begin{table}[htp]
\centering
\scalebox{0.70}{
\begin{tabular}{|c|c|c|c|}
\hline
\textbf{Setting} & \textbf{Training Method} & \textbf{COCO@0.5} & \textbf{COCO@[0.5,0.95]} \\ \hline
\textbf{With training data}            & without KD~\cite{ren2015faster} & 53.8          & 33.9          \\ \hline
\multirow{2}{*}{\textbf{Without training data}} & without KD (\textbf{Ours})   & 30.9           & 15.6           \\ \cline{2-4} 
                                       & with KD (\textbf{Ours})        & \textbf{41.3} & \textbf{24.0} \\ \hline
\end{tabular}
}
\caption{\small{Performance of our proposed method on COCO dataset (w/ and w/o KD) on Resnet-$101$}}
\label{tab:coco}
\end{table}

We also perform experiments on COCO~\cite{lin2014microsoft} which is large scale object detection dataset with $80$ object categories. The models are evaluated as 
: a) average precision 
with IoU at $0.5$ and b) mean of the average precisions calculated with IoU starting with $0.5$ to $0.95$ having a step size of $0.05$. 
We denote the former one as COCO@$0.5$ and later one as COCO@[$.5$,$.95$]. The results are shown in Table~\ref{tab:coco}. We use Resnet-$101$ pretrained model~\cite{jjfaster2rcnn} on COCO as the \Te{} network which obtains $53.8\%$ and $33.9\%$ mAP on evaluation using COCO@$0.5$ and COCO@[$.5$,$.95$]. 
Despite of the large \Te{} architecture and complex training dataset, our zero-shot kd method using our pseudo-dataset as transfer set obtains decent performance of $41.3\%$ and $24.0$\%  while $30.9\%$ and $15.6\%$ without kd on COCO@$0.5$ and COCO@[$.5$,$.95$] 
respectively. Hence, our proposed method is scalable even on large scale detection datasets. 

We also tried to use other well-known losses like TV loss~\cite{rudin1992nonlinear} and L2 regularizer to generate natural-looking MOIs but we observed that such losses 
did not yield improvement in mAP. We also empirically demonstrated that our prepared pseudo-targets reasonably capture the label distribution of training data. Refer supplementary for more details and analysis.
\vspace{-0.1in}
{
\begin{figure}[htp]
\centering
\centerline{\includegraphics[width=0.99\textwidth]{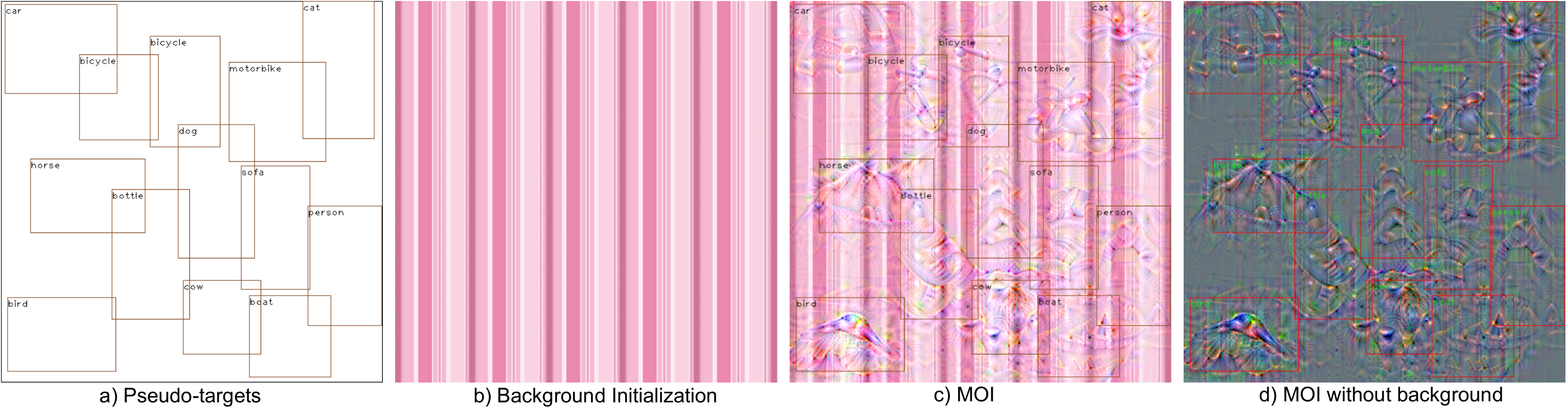}}
\scriptsize
\vspace{-0.1 in}
\caption{\small{Visualization of a synthesized sample: a) Pseudo-Targets using Algo.1, b) Background Init. using texture image, c) MOI obtained using Algo.2, d) MOI without background for better visibility.
}}
\label{fig:visualization}
\end{figure}

\subsection{Efficacy of our pseudo-dataset beyond KD}
\label{sec:beyond_kd}
\vspace{-0.18in}
\begin{table}[htp]
\begin{minipage}[htp]{0.48\textwidth}
  \centering
  \scalebox{0.65}{
\begin{tabular}{|c|c|c|c|}
\hline
\multirow{4}{*}{\textbf{Architecture}} &
  \textbf{Original Dataset ($D$)} &
  \multicolumn{2}{c|}{\textbf{Our Pseudo-dataset ($\hat{D}$)}} \\ \cline{2-4} 
 &
  \begin{tabular}[c]{@{}c@{}}mAP \\ (Upper Bound)\end{tabular} &
  \begin{tabular}[c]{@{}c@{}}mAP\\ (w/ KD)\end{tabular} &
  \begin{tabular}[c]{@{}c@{}}mAP \\ (training \\ from scratch)\end{tabular} \\ \hline \hline
\begin{tabular}[c]{@{}c@{}}Resnet-18\\  (KITTI)\end{tabular}  & 73.3 & 64.2 & \textbf{57.8} \\ \hline
\begin{tabular}[c]{@{}c@{}}Resnet-34 \\ (Pascal)\end{tabular} & 70.1 & 58.6 & \textbf{50.9} \\ \hline
\begin{tabular}[c]{@{}c@{}}Resnet-101\\  (COCO)\end{tabular}  & 53.8 & 41.3 & \textbf{30.9} \\ \hline
\end{tabular}}

\caption{\small{Performance (in \%) on our pseudo-dataset when used to train the network from scratch across architectures and datasets.}}
\label{tab:train_from_scratch}
\end{minipage}\qquad
\begin{minipage}[htp]{0.46\textwidth}
  \centering
  \scalebox{0.68}{
\begin{tabular}{|c|c|c|}
\hline
\textbf{Setting}                     & \textbf{Dataset}  & \textbf{mAP} \\ \hline \hline
\multirow{4}{*}{\textbf{Proxy Data}} & Cityscapes        & 53.1         \\ 
                                     & Cityscapes + \textbf{Ours} & 62.0 ({\color{red}{+ 8.9}})    \\ \cline{2-3} 
                                     & Synthia           & 55.6         \\ 
                                     & Synthia + \textbf{Ours}    & 62.8 ({\color{red}{+ 7.2}})     \\ \hline \hline
\multirow{4}{*}{\textbf{\begin{tabular}[c]{@{}c@{}}Few original \\ training samples\end{tabular}}} & 5\% KITTI & 58.5 \\ 
                                     & 5\% KITTI + \textbf{Ours}  & 64.8 ({\color{red}{+ 6.3}})     \\ \cline{2-3} 
                                     & 10\% KITTI        & 60.1         \\ 
                                     & 10\% KITTI + \textbf{Ours} & 66.2 ({\color{red}{+ 6.1}})     \\ \hline
\end{tabular}
}
\caption{\small{Performance (in \%) on our pseudo-dataset when used as augmentation on Resnet-$18$ without KD}}
\label{tab:as_aug}
\end{minipage}
\end{table}
\vspace{-0.1in}
As shown in Table~\ref{tab:train_from_scratch}, we obtain respectable performances across datasets and architectures when the network is trained from scratch using our pseudo-dataset ($\hat{D}$) without any Teacher assistance. We observe a similar performance gap between the network trained on $\hat{D}$ and $D$, even on a large scale challenging dataset like COCO. This consistent behaviour highlights the scalability of our proposed approach across datasets and architectures. Our pseudo-dataset $\hat{D}$ has a similar behaviour as $D$. For instance; our generated pseudo-samples are robust to augmentations like original data. Also, when KD is applied on $\hat{D}$, we obtain a significant performance improvement similar to the data-KD setup.

We further analyse the efficacy of our pseudo-dataset by investigating its performance under different scenarios: a) proxy data + pseudo-data and b) few samples of training data (few-shot) + pseudo-data. We take an equal number of samples from proxy data and pseudo-data for a fair comparison. Also, total number of samples is similar to the size of the KITTI training dataset. 
The models are evaluated using the official KITTI evaluation tool. As shown in Table~\ref{tab:as_aug}, we obtain a noticeable performance improvement of $\approx 6-9\%$ when our generated data ($\hat{D}$) is used in conjunction with either proxy data or few samples of original training data. However, we do not observe any significant performance improvement when our pseudo-data is used along with all training samples. \textit{The performance on other state-of-the-art object detection approaches such as Yolo~\cite{redmon2018yolov3} and FCOS~\cite{tian2019fcos}, are put in the supplementary (sec. $1$) where our dataset ($\hat{D}$) is used as training set.} Overall, the observations from Tables~\ref{tab:train_from_scratch} and ~\ref{tab:as_aug} and the results from supplementary indicate that our pseudo-dataset ($\hat{D}$) can be treated as reliable representatives of the original training data and can be used in applications beyond KD where such data is required but is present in small amounts or not present at all.}
\vspace{-0.2in}
\section{Conclusion}
\label{sec:conclusion}
\vspace{-0.13in}
Recent approaches have focused much on data-oriented knowledge distillation (KD) for object detection. However, there are limitations on the availability of the training data due to data privacy and sensitivity concerns. To handle them, we proposed a novel zero-shot KD method on two-stage Faster RCNN object detection models. Through extensive experiments on several architectures and datasets, we showed the utility of our generated data as a transfer set in KD. 
We observe decent mAP even when a network is trained on our generated data without \Te{} assistance. 
Our pseudo-dataset further leads to significant improvement in mAP when augmented with proxy data or few training samples, proving them as reliable estimates of original training data. However, our current method depends on the anchor information from the pretrained model to get an estimate about the range of object sizes in the original training dataset. As a future direction, we will reduce this dependency further to make them suitable for anchor-free detectors. Also, extending our work to black box setting, assuming no access to the pretrained model weights would be another interesting direction.

\newpage
\onecolumn
\begin{center}
    \LARGE{\textbf{ {\textit{Supplementary for:} \\``\textit{Beyond Classification:} Knowledge Distillation using Multi-Object Impressions''} }}

\end{center}

\setcounter{section}{0}
\setcounter{table}{0}
\setcounter{figure}{0}
\setcounter{equation}{0}
\vspace{13pt}
\hrule
\vspace{13pt}

\section{Efficacy of our pseudo-dataset across different Object Detection Methods}
\label{sec:compare_detection_approaches}
We investigate the efficacy of our generated pseudo-dataset ($\hat{D}$) obtained from Pascal trained Faster-RCNN network with Resnet-$34$ backbone. We train different state-of-the-art detection models on our pseudo-dataset. Particularly, we evaluate on Faster-RCNN with FPN~\cite{lin2017feature}, YOLO~\cite{redmon2018yolov3}, FCOS~\cite{tian2019fcos} and RetinaNet~\cite{lin2017focal}. 
\vspace{0.3in}
\begin{table}[htp]
\centering
\scalebox{0.8}{
\begin{tabular}{|c|c|c|}
\hline
\multirow{2}{*}{\textbf{Object Detection Methods}} &
  \textbf{\begin{tabular}[c]{@{}c@{}}Original dataset ($D$)\\ (Pascal)\end{tabular}} &
  \textbf{Our Pseudo-dataset ($\hat{D}$)} \\ \cline{2-3} 
 &
  \begin{tabular}[c]{@{}c@{}}mAP \\ (Upper Bound)\end{tabular} &
  \begin{tabular}[c]{@{}c@{}}mAP \\ (training from scratch)\end{tabular} \\ \hline \hline
\textbf{Faster RCNN}~\cite{ren2015faster}                                                                      & 70.1 & 50.9 \\ \hline
\begin{tabular}[c]{@{}c@{}}\textbf{Feature Pyramid Network}~\cite{lin2017feature}\\ \textbf{(Faster-RCNN)}\end{tabular} & 70.7 &  50.6\\ \hline
\textbf{Yolo V3}~\cite{redmon2018yolov3}                                                                          & 69.2 & 41.1 \\ \hline
\textbf{FCOS}~\cite{tian2019fcos}                                                                             & 64.0 & 46.8 \\ \hline
\textbf{RetinaNet}~\cite{lin2017focal}                                                                             & 71.3 & 51.3 \\\hline
\end{tabular}
}
\caption{\small{Performance (in \%) with our generated samples using different state-of-the-art approaches for the object detection task in the complete absence of the Pascal dataset. The models are evaluated using the Pascal VOC convention style.}}
\label{tab:across_detection_methods}
\end{table}
\begin{enumerate}
\item \textbf{Feature Pyramid Network (FPN)} is often used in addition to the Faster-RCNN model to obtain better feature representations for object detection task. It combines features in bottom-up pathway to yield high-resolution features and top-down pathway to yield low-resolution features. Our pseudo-dataset was not synthesized to deal with such a setup. As shown in Table~\ref{tab:across_detection_methods}, we obtain $50.6\%$ mAP which is very close to the network trained on Faster-RCNN. This implies that our pseudo-dataset is even applicable to the networks that perform different modifications at the feature level.
\item \textbf{YOLO}: We use YOLO-v3 for our experiments. Note that, unlike Faster-RCNN, YOLOv3 is a single stage network. As evident in Table~\ref{tab:across_detection_methods}, we achieve $41.1\%$ mAP even though the pseudo-dataset was generated using Faster-RCNN. This implies that our pseudo-dataset is generalizable to one-stage methods.
\item \textbf{FCOS} is a single stage object detector like Yolo. But it predicts on per pixel basis unlike anchor based methods like Yolo. We obtain $46.8\%$ mAP using FCOS.
\item \textbf{RetinaNet} is also a single stage object detector but uses Focal loss i.e. a modified version of the Cross Entropy loss, that helps to overcome the foreground and background class imbalance problem. We obtain decent mAP of $51.3\%$ which shows that our generated data facilitates training of detection model with any additional losses.

\end{enumerate}

Despite methodological and architectural differences, we obtain respectable performances ($\approx 41-51\%$ mAP) in Table~\ref{tab:across_detection_methods} while training the network from scratch with our pseudo-dataset ($\hat{D}$) using different detection methods. These observations emphasize that $\hat{D}$ reasonably estimates the training data distribution and is applicable to different object detection methods in the absence of training data. \\
\vspace{-0.25in}
\section{Result Analysis and Discussions}
\label{sec:discussion}
In this section, we briefly analyze the results and discuss our observations that we found based on the experiments across architectures and datasets. 
\\
\\
\textbf{Convergence}: Unlike existing data-free works such as~\cite{nayak2019zero, Yin_2020_CVPR} takes large iterations and longer time for generation. On the contrary, each batch in proposed Algorithm $2$ in the main draft converges within $100$ iterations for even large scale datasets like COCO.
\\
\\
\textbf{Other Losses}: In order to further improve the distillation performance, we even tried to use other well-known natural image priors like Total variation~\cite{rudin1992nonlinear} and L2 regularizer to generate natural-looking Multi-Object Impressions (MOIs). 
But we observed that such losses though are helpful in improving visualization of samples but do not yield improvement in mAP. Sometimes, it even leads to lower performance during distillation. We also observed that learning rate (lr) is a crucial hyperparameter whose careful tuning can lead to the generation of a better transfer set. Infact, we found that only tuning of lr works better than using such additional losses. Please refer to section~\ref{sec:other_losses} for experimental results and more details.
\\
\\
\textbf{Beyond transfer set}: In order to further study how well our pseudo-dataset has captured the training data distribution, we use our generated data to train the network from scratch. From Table $1$ in the main draft, we can observe that we obtain a decent performance of $57.8\%$ mAP on KITTI, even when Resnet-$18$ is trained from scratch using MOIs. Moreover, we obtained similar respectable performances even in the case of complex and large scale datasets like Pascal and COCO. In the case of Pascal, (as shown in Table $2$ in main draft) VGG-$16$ and Resnet-$34$ networks obtains $49.3\%$ and $50.9\%$ mAP while training them with our MOIs with no \Te{} assistance. Similarly, (Table $3$ in main draft) with no KD, we observed reasonable performance of $30.9\%$ 
mAP on evaluation using COCO@0.5. 
{The summarized results across datasets and architectures while training the network from scratch with our generated datset ($\hat{D}$) is put in Table 4 in the main draft. Also, our pseudo-dataset is even suitable to be used as augmentation when arbitrary data or few training samples are present (shown in Table 5 in main draft). Moreover, we obtain reasonable performances across different detection methods (Sec.~\ref{sec:compare_detection_approaches}). Thus, our novel pseudo-dataset synthesizing framework can have a lot of practical utility.}
These observations show that our generated data has even the potential to be used beyond a transfer set for distillation, such as training a network from scratch. Such insights are important and currently missing in existing data-free methods which requires further investigation.
\\
\\
\textbf{Performance Analysis}: We analyze both the stages of our generation strategy : generation of pseudo-targets and generation of MOIs. 

To analyze the first stage, we replace our pseudo-targets with original training data ground truths. In other words,  we use label information from the training data like the position, size, and class label of an object rather than actual objects in the images and then generate impressions corresponding to them. We observed only a minor gain in performance, which shows our prepared pseudo-targets reasonably capture the label distribution of training data. More details are in section~\ref{sec:dist_targets}.

 Next, we analyze stage two which deals with generation of MOIs. Since neither the training data nor its statistics are assumed to be available, there always exists a domain gap between our generated data and original training data. We tried to reduce this gap by better initialization i.e. initializing MOIs with textures rather than random initialization. 
 The original training data is independent of the architecture. But our generated MOIs depend on the \Te{} architecture. 
 To decouple the MOIs from the \Te{}, we use differential batch augmentation which makes them more robust and hence results in improved performance. 
 \vspace{-0.1in}
\section{Motivation for Power Law distribution}
\label{power_law}
As discussed in Algorithm 1 (in the main draft), the maximum number of possible objects in any sample is denoted by $M$. While preparing the target labels,  the number of objects for each sample (denoted by $N_{i}$ for $i^{th}$ sample)  is uniformly sampled between $1$ to $M$. For simplicity, we restrict our discussion of objects’ size with respect to a particular sample and use $N$ to denote the number of objects belonging to that sample. The minimum and maximum possible object sizes obtained using the information of anchor scales are denoted by $A^{min}$ and $A^{max}$ respectively. This implies that the sizes of each object need to lie in the range [$A^{min}, A^{max}$]. 
\\
\\
\textbf{Uniform Sampling ($\mathcal{U}(A^{min}, A^{max})$)}: An intuitive strategy for deciding objects' size is to uniformly sample each object size from the range [$A^{min}, A^{max}$]. If one of the sampled object sizes is very large (close to $A^{max}$), then it would result in high overlaps while placing other objects. In extreme cases when $N$ is large, it may end up with most of the objects contained inside another object. Therefore, with such a strategy it will be difficult to satisfy the $IoU_{threshold}$ constraint. The major reason for high overlaps is that there is no restriction on the object sizes and each object can be as large as $A^{max}$. In order to overcome this problem, we need to enforce a constraint on $A^{max}$.
\\
\\
\textbf{Constrained Uniform Sampling ($\mathcal{U}(A^{min}, min((W\cdot H)/N, A^{max}))$)}: One way to enforce the constraint on $A^{max}$ is to have maximum object size as $A^{{'}_{max}} \leftarrow min((W\cdot H/N), A^{max})$ which is fixed for all the objects in a sample. $W$ and $H$ denote the width and height of the sample respectively. Since each object’s maximum size is same and reduced to $A^{{'}_{max}}$,  this would help in placing the objects within the $IoU_{threshold}$ constraint. By putting such restrictions on the maximum object size where $A^{{'}_{max}}$ becomes small with a large value of $N$, objects of small sizes are more favoured.
\\
\\
\textbf{Interval based Uniform Sampling}: In order to explicitly enforce the favoring of large object sizes and reduce biases towards the small sizes, we can adopt a interval based uniform sampling strategy ($\mathcal{U}(max(A^{min},(W\cdot H)/(N+1)), min(W\cdot H/(N), A^{max})$). For e.g., when $N$ is $1$, the interval  
can be used to allow objects of large sizes. Similarly, for $N=2$, the object sizes can be sampled in the interval [$max(A^{min},(W\cdot H)/3), min((W\cdot H)/2, A^{max})$] 
and so on. We can perform uniform sampling on each of these intervals. The problem with this strategy is that the objects are sampled within a particular interval for a fixed value of $N$ and the probability of objects belonging to another interval for the same value of $N$ is zero. That means the small and large sized objects cannot be sampled together for a particular value of $N$. The power law overcomes this problem by allowing objects of small and large sizes to occur together with some probability.
\\
\\
The power law is defined as:
\begin{equation}
P(x; a_{o}) = a_{o}x^{a_{o}-1}, 0 \le x \le 1, a_{o}>0
\end{equation}
Let sampling from power law be represented by : $x \sim P(a_{o})$.
The mean of the power law distribution is given by $\dfrac{a_{o}}{a_{o}+1}$.

\begin{figure}[htp]
\centering
\centerline{\includegraphics[width=0.5\textwidth]{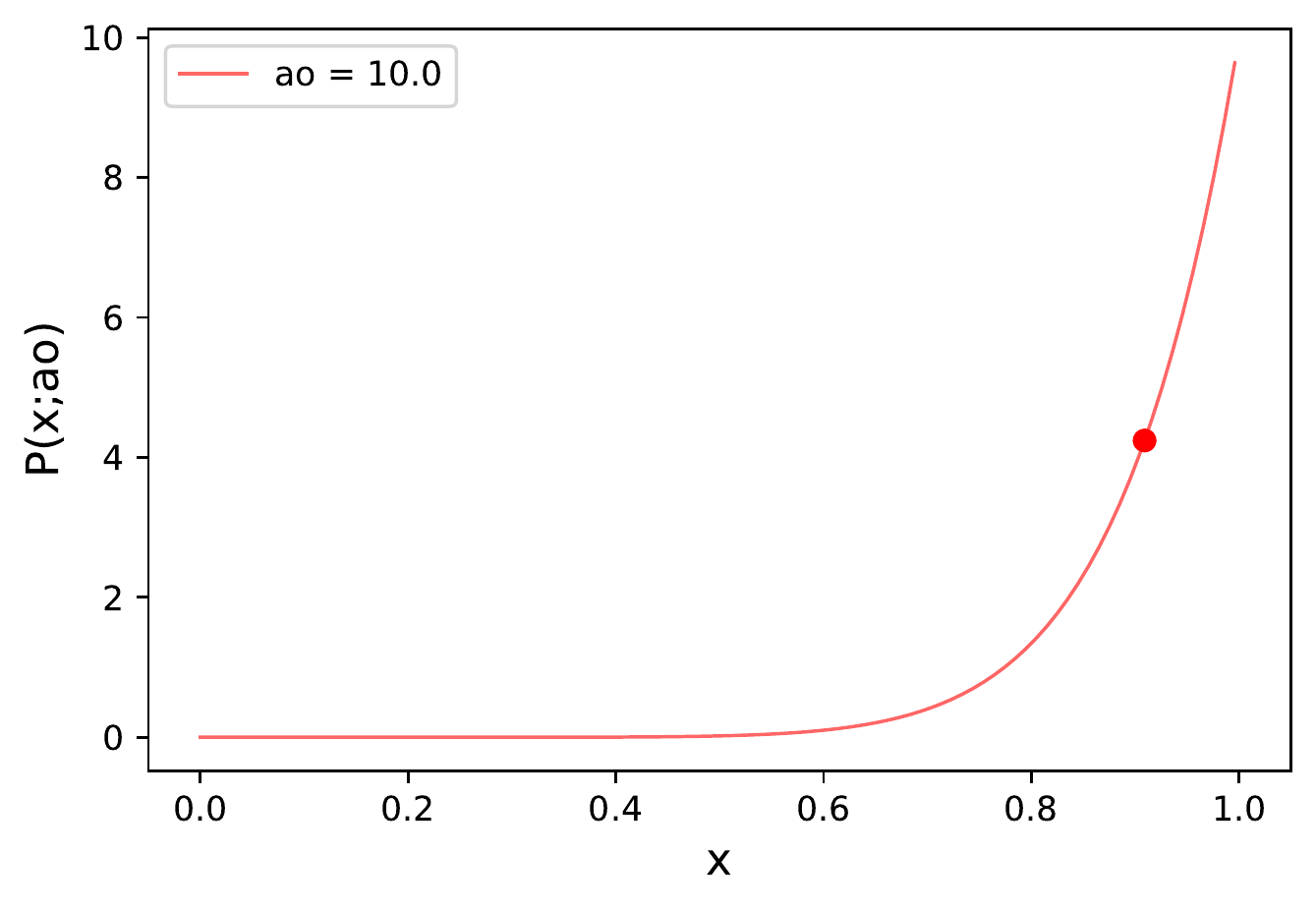}}
\caption{\small{The Power Law distribution with $a_{o}=10$. $x=0$ represents $A^{min}$ and $x=1$ represents $A^{{'}_{max}}$. The red dot represents the mean of the distribution which is favouring areas closer to $A^{{'}_{max}}$.}}
\label{fig:power_law_fixed}
\end{figure}

\textbf{Power Law with fixed $a_{o}$} : The power law distribution defined using $a_{o}$ is used to sample $x$ which lies between $0$ to $1$.  The sampled $x$ is rescaled to interval [$A^{min}, A^{{'}_{max}}$] where $A^{{'}_{max}}$ is $min((W\cdot H)/N, A^{max})$]. We assume that $a_{o}$ is fixed. This would imply that the mean of the distribution is also fixed. Thus, the majority of the sampled objects' sizes lie around the mean (as shown in Figure~\ref{fig:power_law_fixed}. When $a_{o}=1$, it is same as constrained uniform sampling which favours small areas. While on the other hand, for a large value of $a_{o}$, mean is very close to $1$ which leads to negligence of small areas. This observation motivates to have a variable $a_{o}$.
\\
\begin{figure}[htp]
\centering
\centerline{\includegraphics[width=0.5\textwidth]{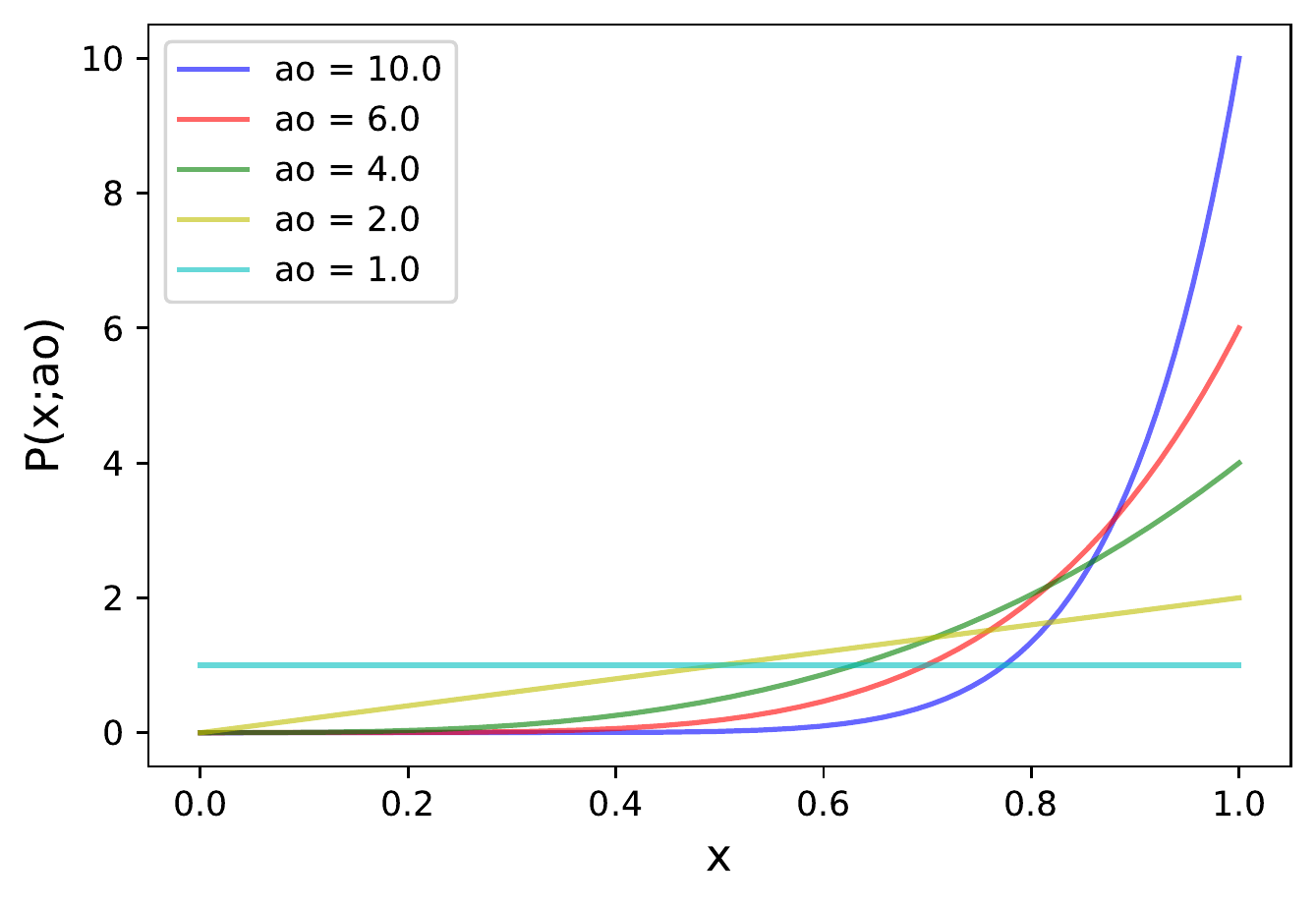}}
\caption{Power Law distribution for different values of $a_{o}$. As the value of $a_{o}$ decreases to $1$, the distribution gradually switches to Uniform distribution.}
\label{fig:power_law_variable}
\end{figure}

\textbf{Power Law with variable $a_{o}$}: Our proposed Algorithm 1 (in the main draft) defines $a_{o}$ as the ratio $\dfrac{M}{N}$ for a particular sample. For different values of $a_{o}$, the distribution curves is shown in Figure~\ref{fig:power_law_variable}. 
For $N=1$, $a_{o} = M$ which allows sampling large objects' sizes with high probability and small objects sizes' with low probability. 
As the value of $N$ increases, we prefer sampling of small sized objects.
When $N=M$, $a_{o}=1$, it is same as constrained uniform sampling. Since it overcomes the limitations of previous strategies, therefore, we use our proposed, variable $a_{o}$ based power law distribution in Algorithm 1 (in the main draft) to define the object sizes for the target labels.

\vspace{-0.15in}
\section{Distribution of objects in prepared Pseudo-Targets}
\label{sec:dist_targets}
\vspace{-0.05in}
Comparison of prepared pseudo-targets and Pascal labels with respect to the distribution of the object sizes is shown below:

\begin{figure}[htp]%
\centering
\begin{minipage}[htp]{0.45\linewidth}
\centering
\centerline{{\includegraphics[width=6cm]{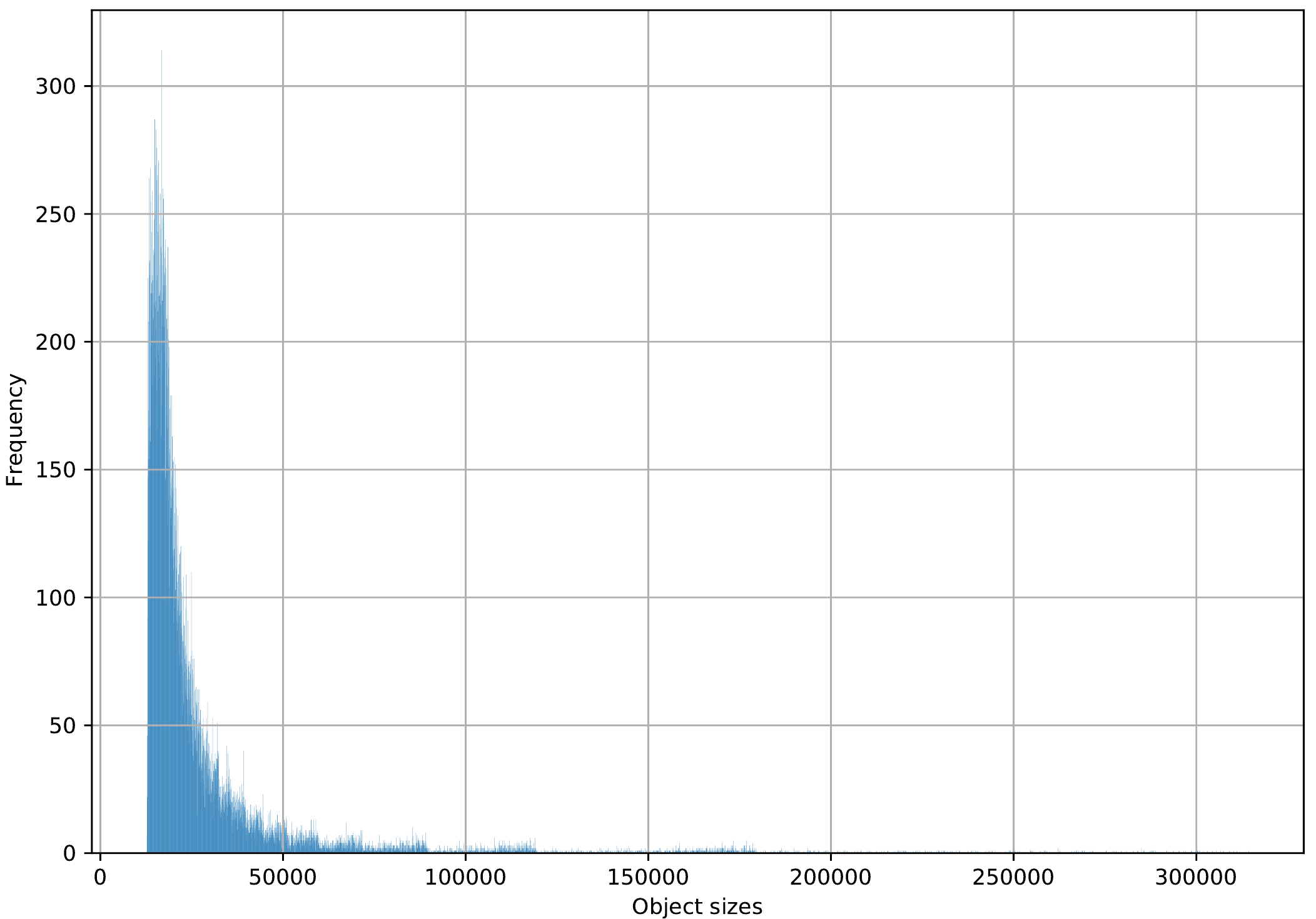} }}
\caption{Histogram plot of object sizes and their frequency on our prepared pseudo-targets.}
\label{fig:hist_ours}
\end{minipage}
\quad
\hspace{0.01\linewidth}%
\begin{minipage}[htp]{0.45\linewidth}
\centering
\centerline{{\includegraphics[width=6.0cm]{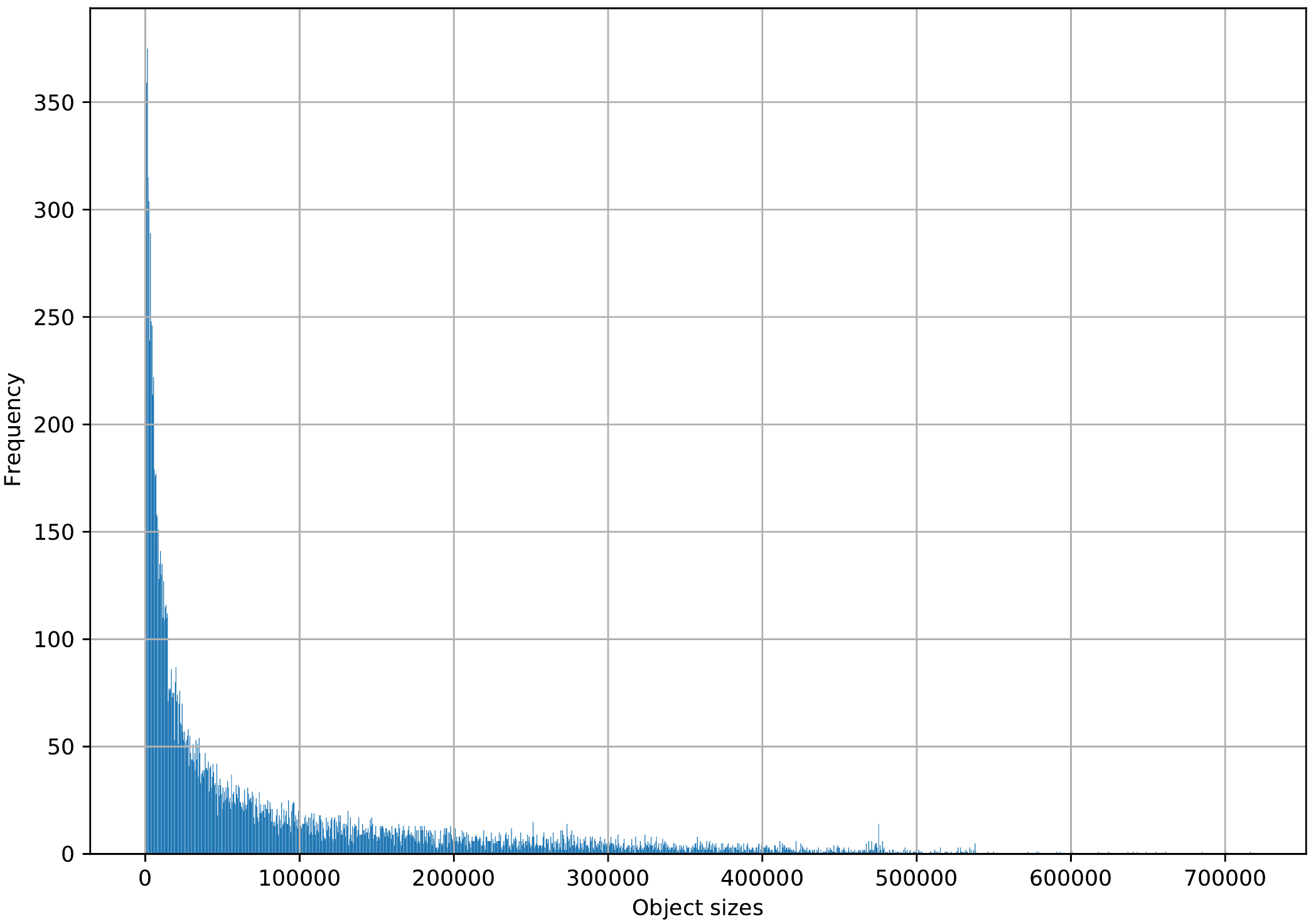} }}
\caption{Histogram plot of object sizes and their frequency from the Pascal Dataset}
\label{fig:hist_original}
\end{minipage}
\end{figure}

From Figure~\ref{fig:hist_ours} and ~\ref{fig:hist_original}, we can observe that in the absence of any prior knowledge of the training data, our pseudo-data obtained using Algorithm 1 (in the main draft) is a good approximate for training label distribution. More precisely, the distribution of object sizes of pseudo-targets using the Power Law distribution and anchor information (ratios and scales) of the \Te{} network reasonably captures the training data distribution of object sizes. However, we do observe 
that we have low count of large sized objects. This may be due to the fact that our MOIs are of fixed $600 \times 600$ dimension. 
Next, we empirically investigate the effect of prior knowledge of the training label distribution on the distillation performance of the \St{} model.

%
\begin{table}[htp]
\centering
\begin{tabular}{|l|c|c|}
\hline
\multirow{1}{*}{Zero-Shot Distillation} &\multicolumn{1}{c|}{Without AI (mAP in \%)} & \multicolumn{1}{c|}{With AI (mAP in \%)} \\ \hline \hline
\multicolumn{1}{|c|}{Resnet 34 $\rightarrow$ Resnet 34} &                                    54.88         &              55.11                            \\ \hline
\end{tabular}
\caption{Distillation using our pseudo-dataset where MOIs are generated without and with additional information (AI). AI is the prior knowledge about the class label, size and location of the objects in the samples of training data (Pascal). Note that the number of samples generated is equal to number of samples in the Pascal training dataset for which the prior label information is available. Thus, $K$ is taken as $5011$ for both with and without AI experiments.}
\label{tab:prior_pascal}
\end{table}

\newpage
As shown in Table~\ref{tab:prior_pascal} , we gain only slight improvement in distillation performance even if the class label, object sizes and their locations on training data are known apriori. This shows that our prepared pseudo-targets reasonably estimates the label distribution of the training data (when only the pretrained model and not the training data is available).  
\vspace{-0.15in}
\section{Experiments with other losses for crafting MOIs}
\label{sec:other_losses}
\textbf{Mask Total Variation Loss ($L_{mtv}$}): If we directly apply the total variation loss~\cite{rudin1992nonlinear} on MOIs, it can also blur out the texture background which is not desired. So, we apply it on the mask which is the difference between the current optimized MOI and initialized MOI. This loss makes the generation of MOIs less sensitive to learning rate and optimization easier by retaining the smoothness. But we do not observe any improvement in mAP while performing distillation from Resnet-$18$ \Te{} trained on KITTI to Resnet-$18$-half with $2500$ generated samples using this additional loss (shown in Table~\ref{tab:other_losses_1}). By only carefully tuning the learning rate (lr) without using this loss, we observe better performance. However, this loss can give improvement in mAP in cases when the lr is set high (shown in Table~\ref{tab:other_losses_2}). But we can clearly observe from Table~\ref{tab:other_losses_1} and~\ref{tab:other_losses_2}, that MOIs synthesized with only $L_{gt}$ loss and no $L_{mtv}$ loss yields better performance when tuned with proper lr.   
\\
\\
\begin{minipage}[b]{.4\textwidth}
  \centering
  \begin{tabular}{|l|c|}
\cline{1-2}
Loss           & mAP (in \%)   \\ \cline{1-2} \hline\hline
$L_{gt}$            & \textbf{27.2}  \\ \cline{1-2}
$L_{gt} + L_{mtv}$     & 20.9   \\ \cline{1-2}
$L_{gt} + 0.1 \cdot L_{mtv}$ & 24.7   \\ \cline{1-2}
\end{tabular}
  \captionof{table}{Distillation when MOIs generated with lr 0.01}
  \label{tab:other_losses_1}
\end{minipage}\qquad
\hspace{0.1\textwidth}
\begin{minipage}[b]{.4\textwidth}
  \centering
  \begin{tabular}{|l|c|}
\cline{1-2}
Loss       & mAP (in \%)  \\ \cline{1-2} \hline \hline
$L_{gt}$         & 15.5  \\ \cline{1-2}
$L_{gt} + L_{mtv}$  & 20.5  \\ \cline{1-2}
\end{tabular}
  \captionof{table}{Distillation when MOIs generated with lr 0.1}
  \label{tab:other_losses_2}
\end{minipage}

\vspace{-0.15in}

\section{Visualization of Multi-Object Impressions (MOIs)}
\label{visual}

Some of the generated samples obtained using Resnet-34 \Te{} trained on Pascal dataset are shown below. 
For each sample we have shown the pseduo-targets, background texture, MOIs with and without background. 
We use additional regularization to highlight the foreground regions for better visualization.
Even though no explicit loss like batch norm is used to enforce the generated samples to look similar to original training data, yet surprisingly many of the patterns in the generated samples can be recognized. The model generates features such as the head of horse, whiskers and face of the cat, fur of the sheep, etc. The model reconstructs the object features on which it paid attention to while training with original training data.
\pagebreak

\begin{longtable}{cc}
\centering

 \vspace{0.2in}
 
\textbf{\large{Pseudo-Targets}} & \textbf{\large{Background Initialization}}
\\
\\
         \begin{minipage}{.45\textwidth}
      \includegraphics[width=\linewidth]{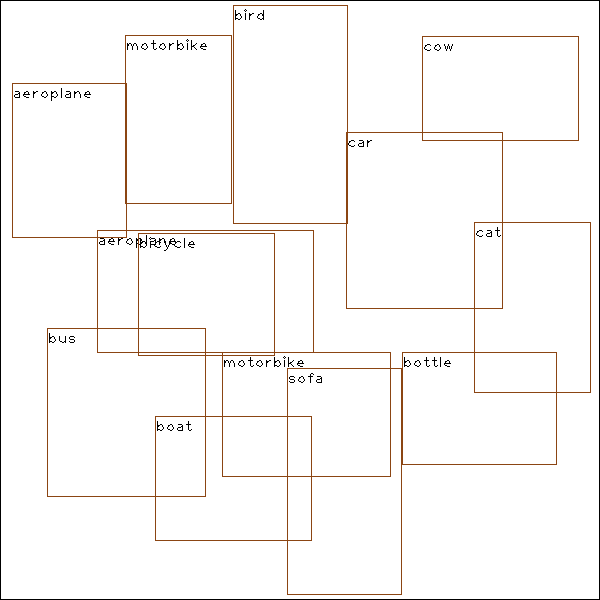}
    \end{minipage}  &               \begin{minipage}{.45\textwidth}
      \includegraphics[width=\linewidth]{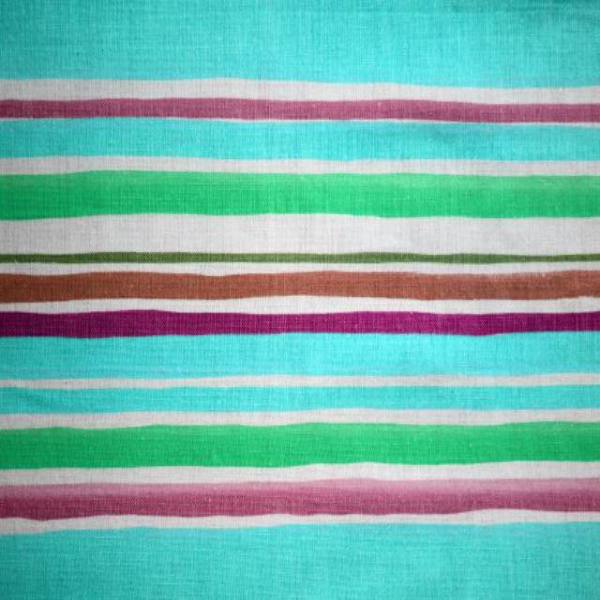}
    \end{minipage}                          \\
    \vspace{0.3in}
    & \\
    & \\
    \vspace{0.2in}
    \textbf{\large{MOI}} & \textbf{\large{MOI without background}}
    \\
    \\
                              \begin{minipage}{0.45\textwidth}
      \includegraphics[width=\linewidth]{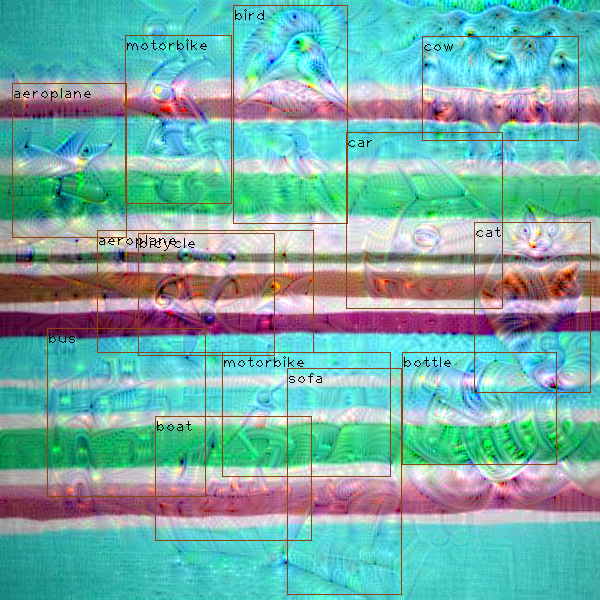}
    \end{minipage}                       &      \begin{minipage}{.45\textwidth}
      \includegraphics[width=\linewidth]{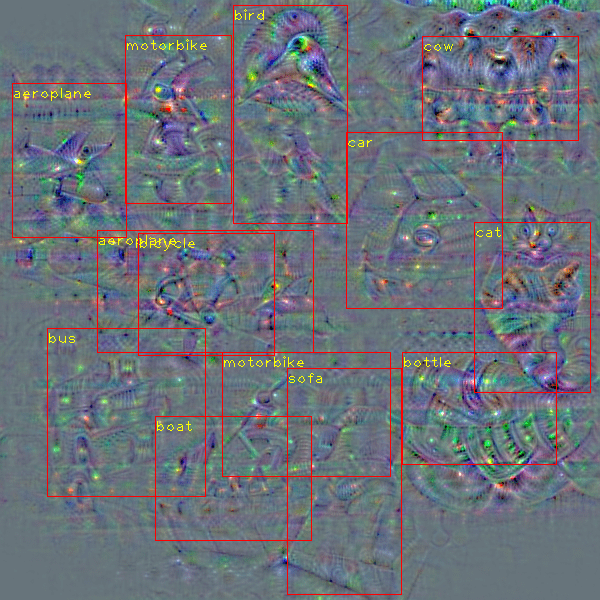}
    \end{minipage}     
    \\
     \pagebreak
     \vspace{0.2in}
              \textbf{\large{Pseudo-Targets}} & \textbf{\large{Background Initialization}}
\\
\\

         \begin{minipage}{.45\textwidth}
         
      \includegraphics[width=\linewidth]{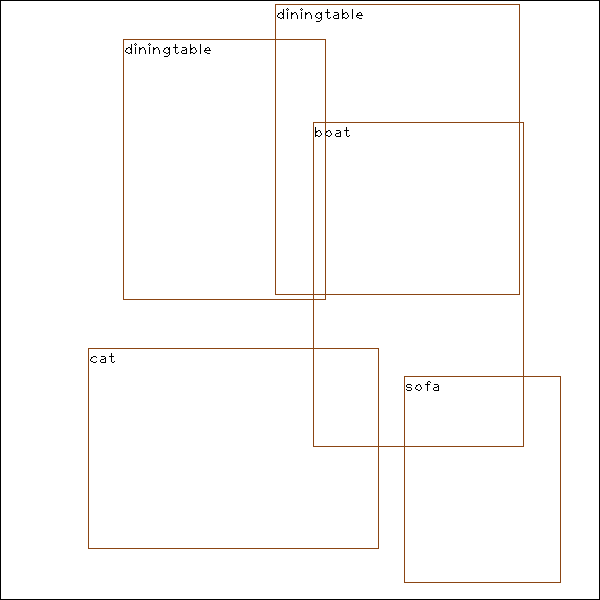}
    \end{minipage}  &               \begin{minipage}{.45\textwidth}
      \includegraphics[width=\linewidth]{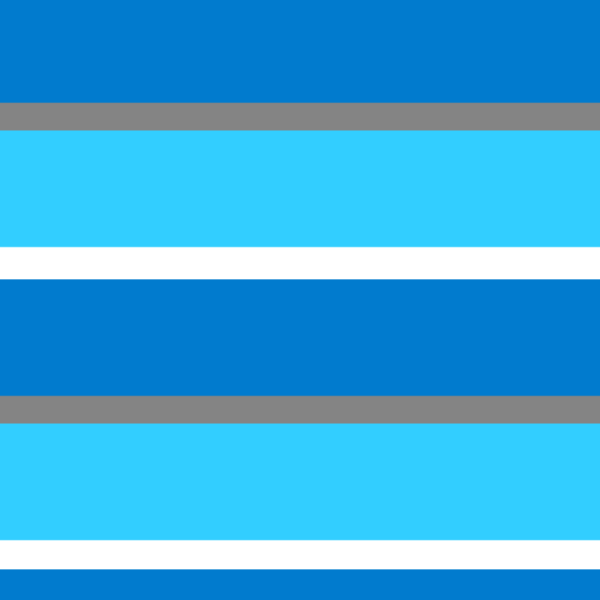}
    \end{minipage}                          \\
    \vspace{0.3in}
    & \\
    & \\
    \vspace{0.2in}
    \textbf{\large{MOI}} & \textbf{\large{MOI without background}}
    \\
    \\
                              \begin{minipage}{0.45\textwidth}
      \includegraphics[width=\linewidth]{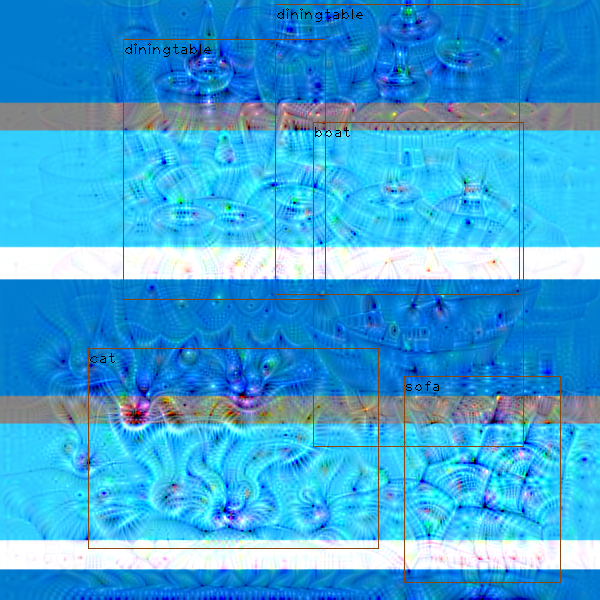}
    \end{minipage}                       &      \begin{minipage}{.45\textwidth}
      \includegraphics[width=\linewidth]{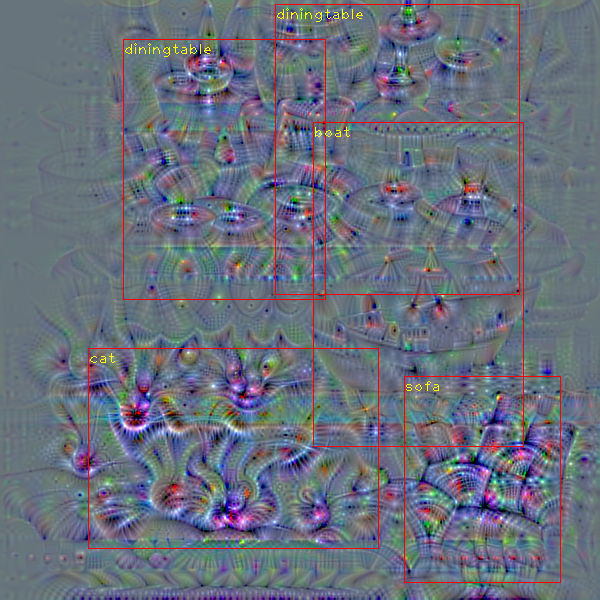}
    \end{minipage}     
    \\
    
    \pagebreak
    \vspace{0.2in}
    \textbf{\large{Pseudo-Targets}} & \textbf{\large{Background Initialization}}
\\
\\
         \begin{minipage}{.45\textwidth}
      \includegraphics[width=\linewidth]{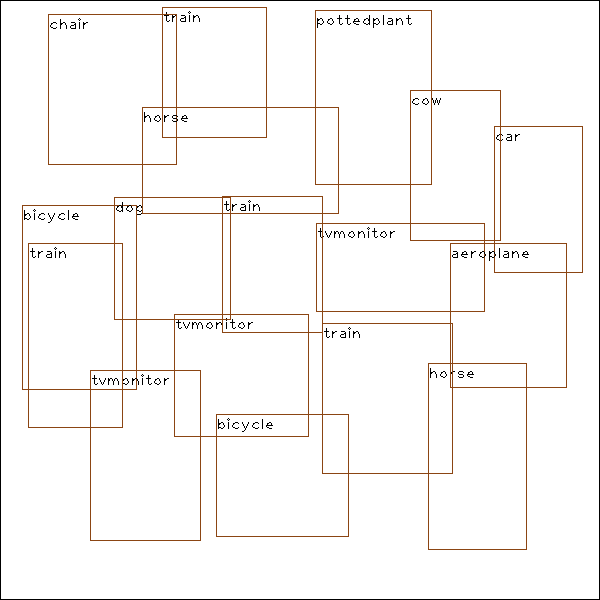}
    \end{minipage}  &               \begin{minipage}{.45\textwidth}
      \includegraphics[width=\linewidth]{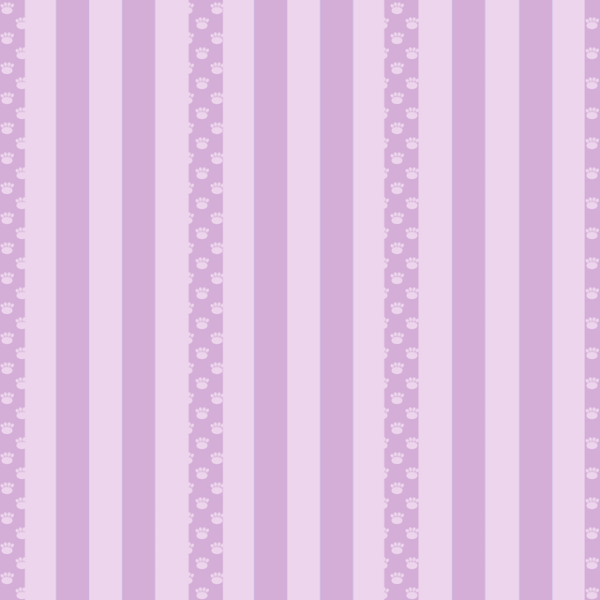}
    \end{minipage}                          \\
    \vspace{0.3in}
    & \\
    & \\
    \vspace{0.2in}
    \textbf{\large{MOI}} & \textbf{\large{MOI without background}}
    \\
    \\
                              \begin{minipage}{0.45\textwidth}
      \includegraphics[width=\linewidth]{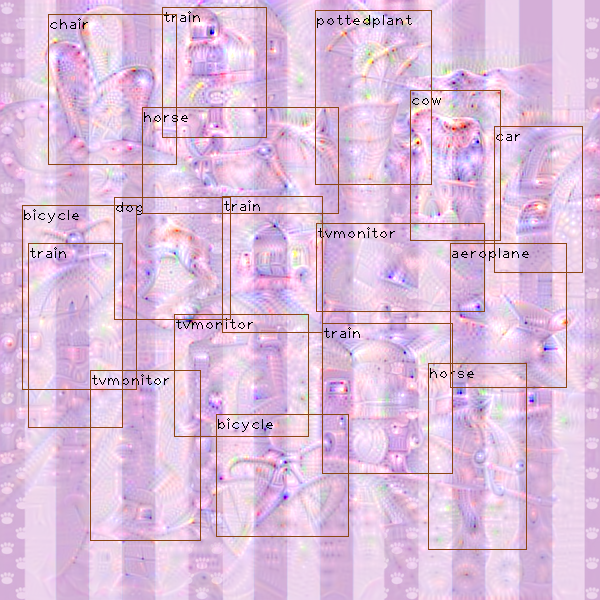}
    \end{minipage}                       &      \begin{minipage}{.45\textwidth}
      \includegraphics[width=\linewidth]{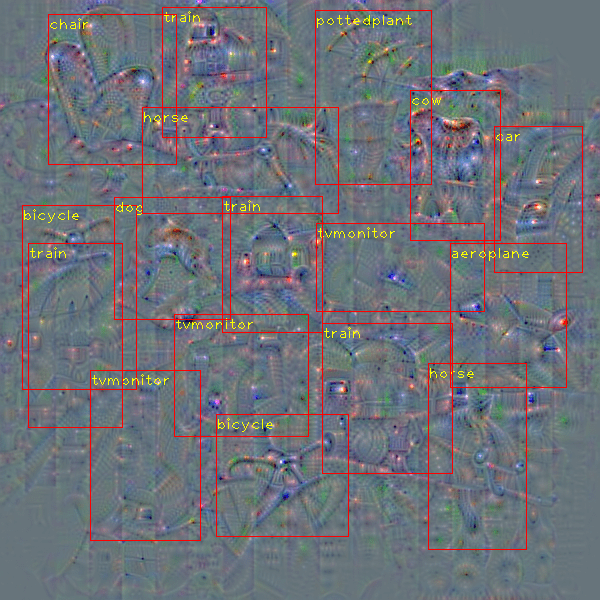}
    \end{minipage}     
    \\
    
    \pagebreak
    \vspace{0.2in}
    \textbf{\large{Pseudo-Targets}} & \textbf{\large{Background Initialization}}
\\
\\
         \begin{minipage}{.45\textwidth}
      \includegraphics[width=\linewidth]{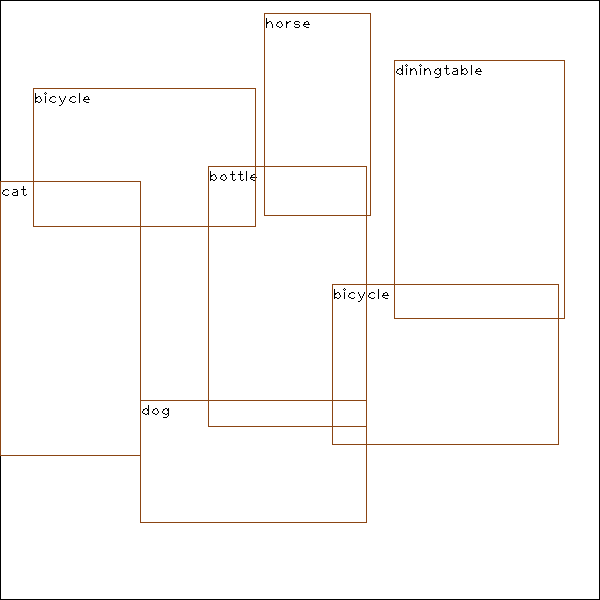}
    \end{minipage}  &               \begin{minipage}{.45\textwidth}
      \includegraphics[width=\linewidth]{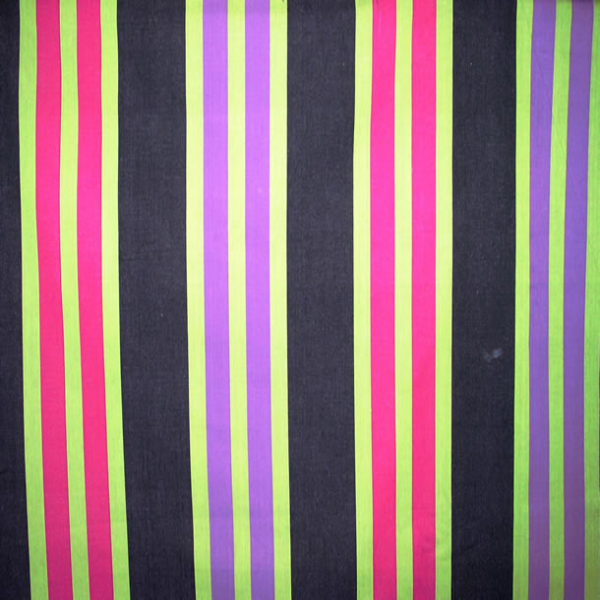}
    \end{minipage}                          \\
    \vspace{0.3in}
    & \\
    & \\
    \vspace{0.2in}
    \textbf{\large{MOI}} & \textbf{\large{MOI without background}}
    \\
    \\
                              \begin{minipage}{0.45\textwidth}
      \includegraphics[width=\linewidth]{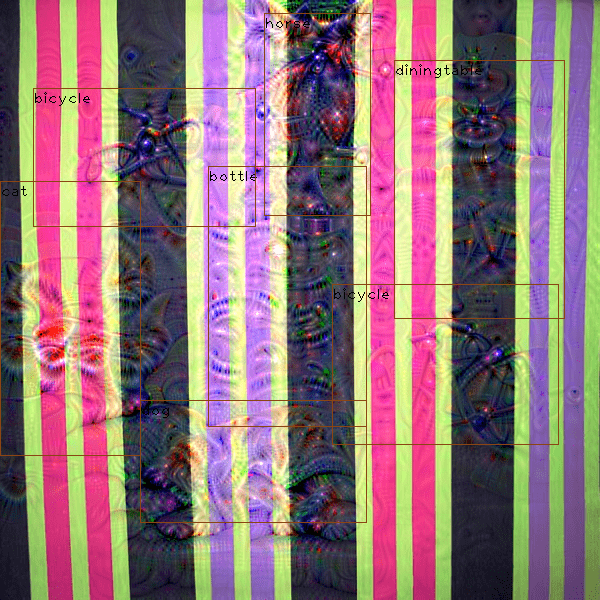}
    \end{minipage}                       &      \begin{minipage}{.45\textwidth}
      \includegraphics[width=\linewidth]{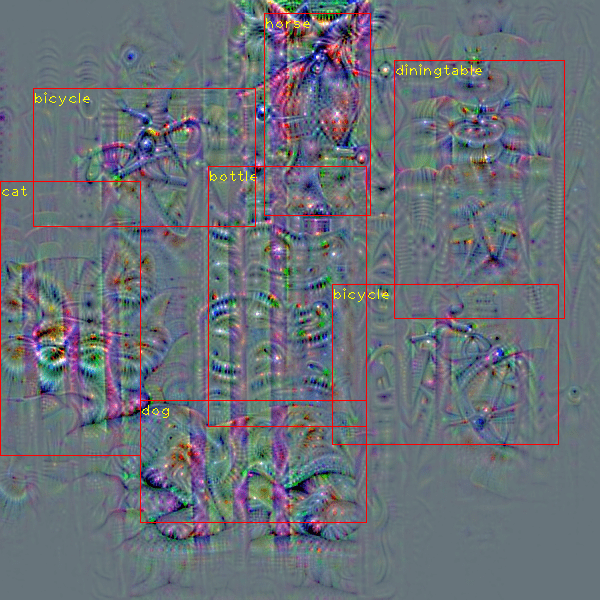}
    \end{minipage}     
    \\
    \pagebreak
    \vspace{0.2in}
    \textbf{\large{Pseudo-Targets}} & \textbf{\large{Background Initialization}}
\\
\\
         \begin{minipage}{.45\textwidth}
      \includegraphics[width=\linewidth]{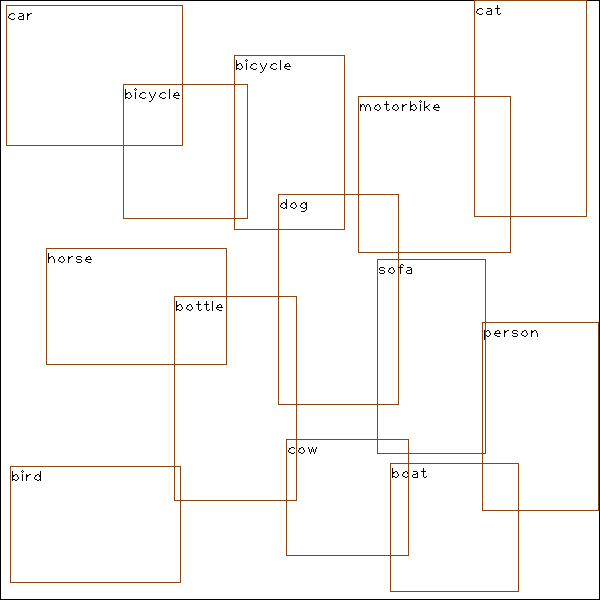}
    \end{minipage}  &               \begin{minipage}{.45\textwidth}
      \includegraphics[width=\linewidth]{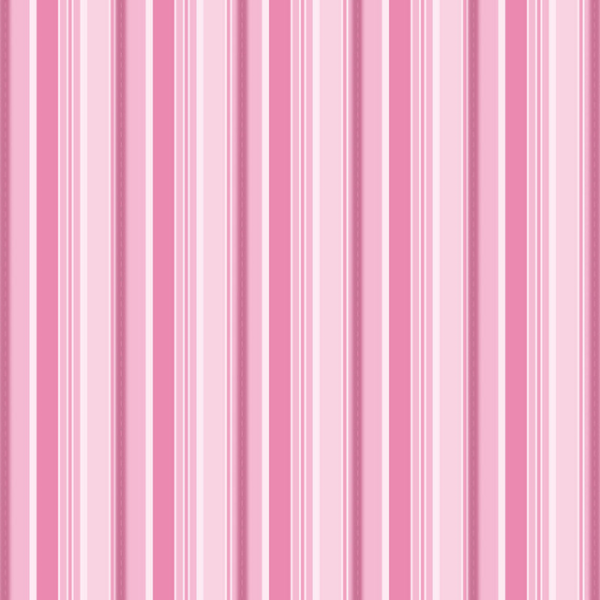}
    \end{minipage}                          \\
    \vspace{0.3in}
    & \\
    & \\
    \vspace{0.2in}
    \textbf{\large{MOI}} & \textbf{\large{MOI without background}}
    \\
    \\
                              \begin{minipage}{0.45\textwidth}
      \includegraphics[width=\linewidth]{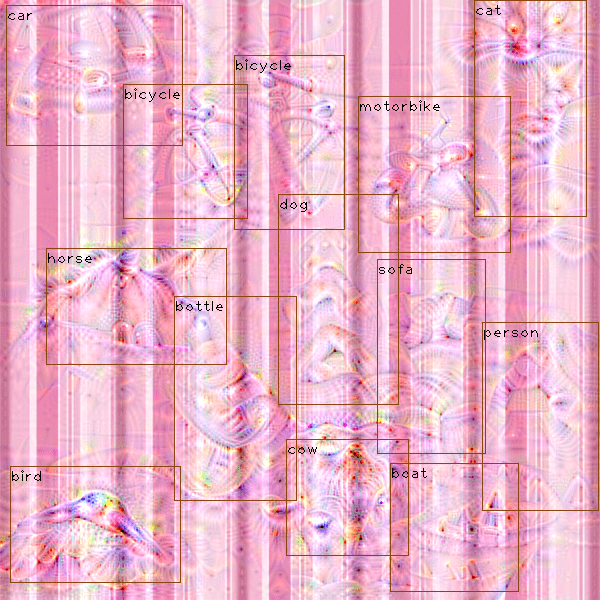}
    \end{minipage}                       &      \begin{minipage}{.45\textwidth}
      \includegraphics[width=\linewidth]{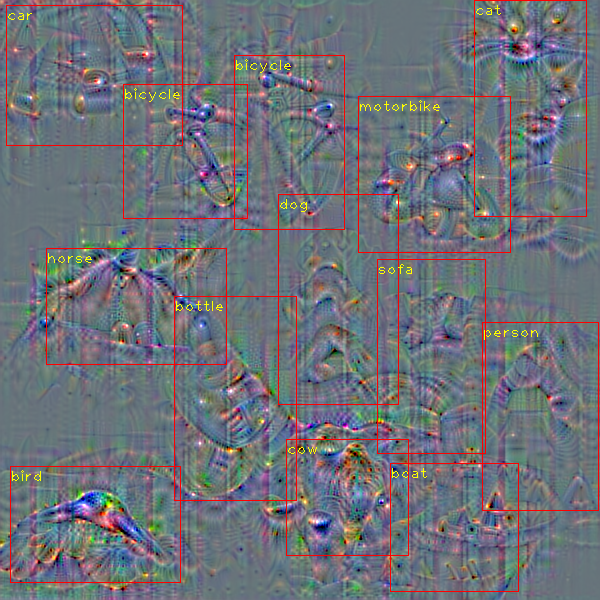}
    \end{minipage}     
    \\
    \pagebreak
    \vspace{0.2in}
    \textbf{\large{Pseudo-Targets}} & \textbf{\large{Background Initialization}}
\\
\\
         \begin{minipage}{.45\textwidth}
      \includegraphics[width=\linewidth]{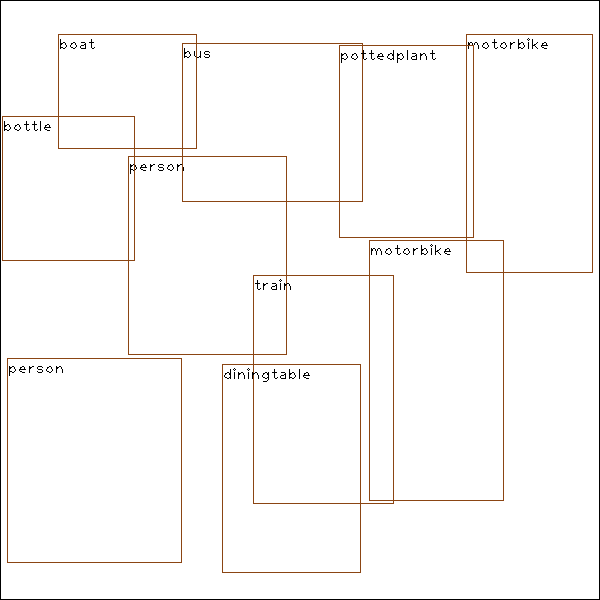}
    \end{minipage}  &               \begin{minipage}{.45\textwidth}
      \includegraphics[width=\linewidth]{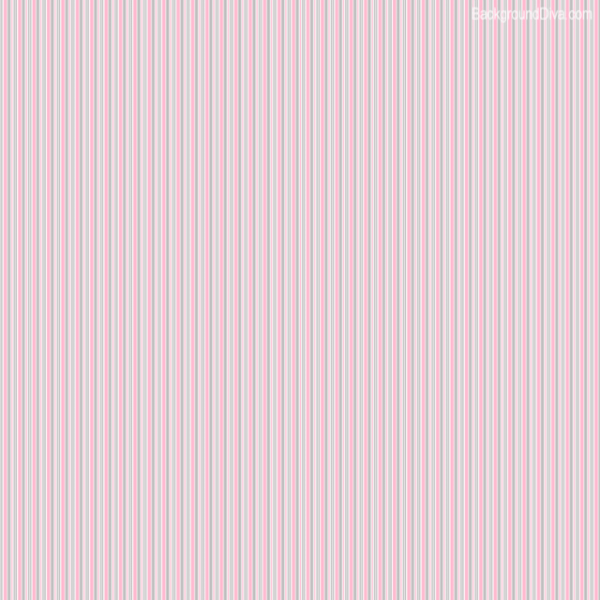}
    \end{minipage}                          \\
    \vspace{0.3in}
    & \\
    & \\
    \vspace{0.2in}
    \textbf{\large{MOI}} & \textbf{\large{MOI without background}}
    \\
    \\
                              \begin{minipage}{0.45\textwidth}
      \includegraphics[width=\linewidth]{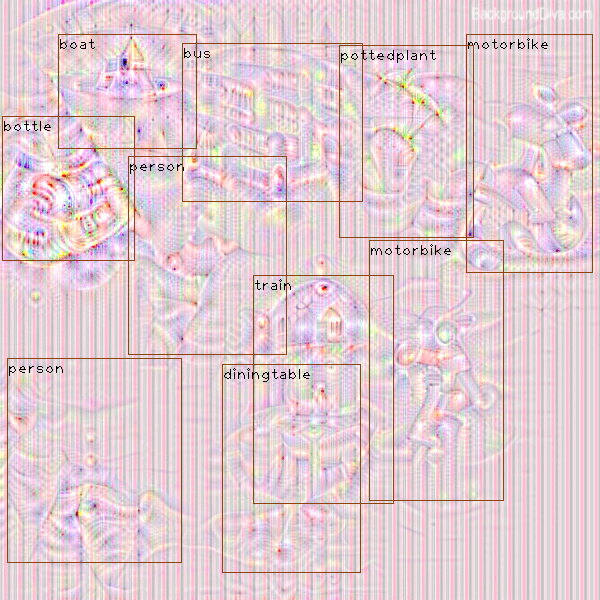}
    \end{minipage}                       &      \begin{minipage}{.45\textwidth}
      \includegraphics[width=\linewidth]{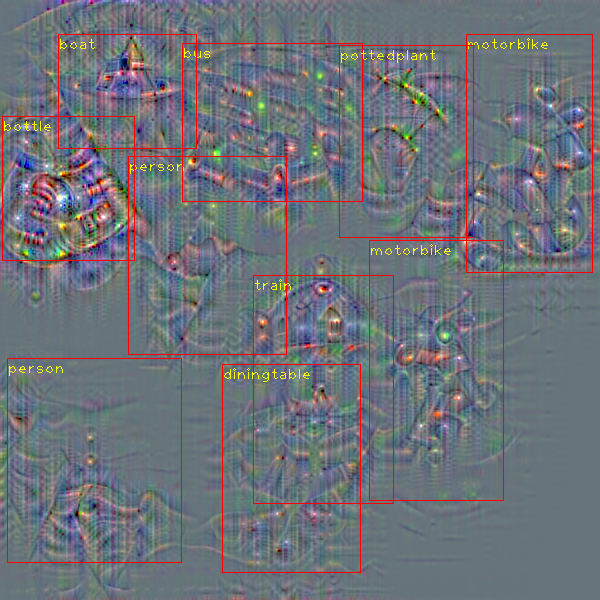}
    \end{minipage}     
    \\
    \pagebreak
    \vspace{0.2in}
    \textbf{\large{Pseudo-Targets}} & \textbf{\large{Background Initialization}}
\\
\\
         \begin{minipage}{.45\textwidth}
      \includegraphics[width=\linewidth]{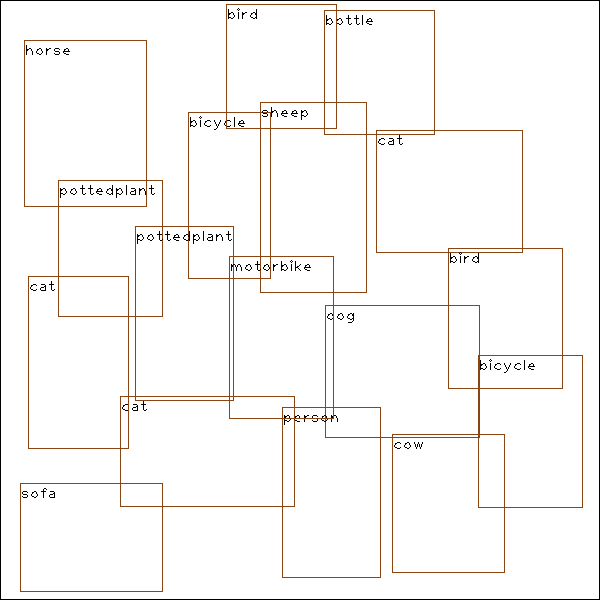}
    \end{minipage}  &               \begin{minipage}{.45\textwidth}
      \includegraphics[width=\linewidth]{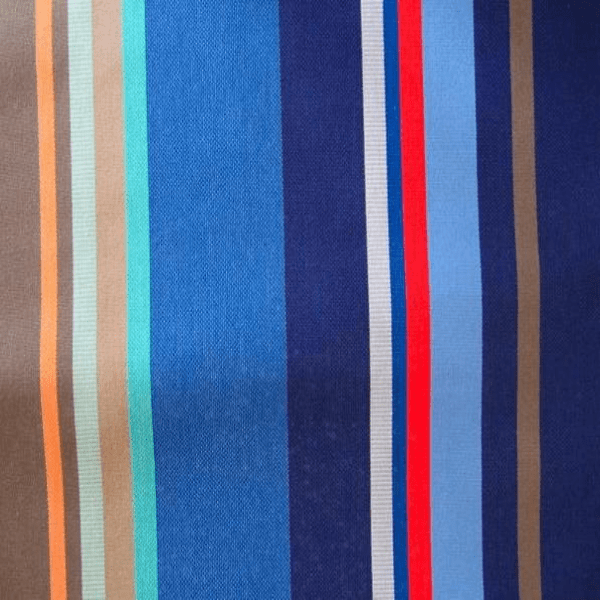}
    \end{minipage}                          \\
    \vspace{0.1in}
    & \\
    & \\
    \vspace{0.2in}
    \textbf{\large{MOI}} & \textbf{\large{MOI without background}}
    \\
    \\
                              \begin{minipage}{0.45\textwidth}
      \includegraphics[width=\linewidth]{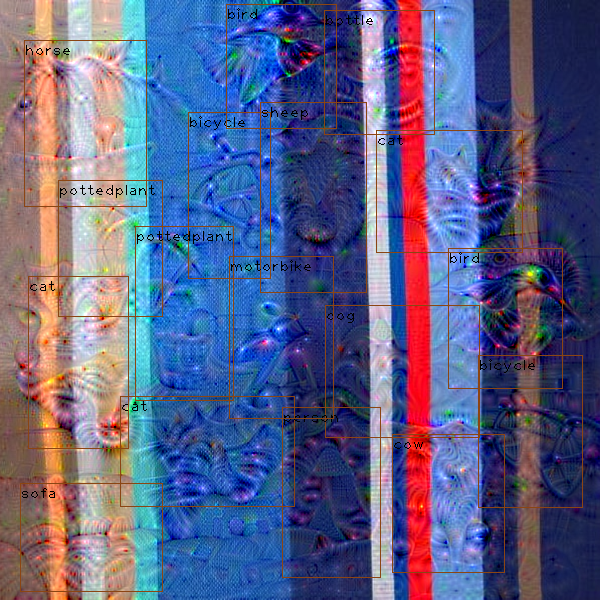}
    \end{minipage}                       &      \begin{minipage}{.45\textwidth}
      \includegraphics[width=\linewidth]{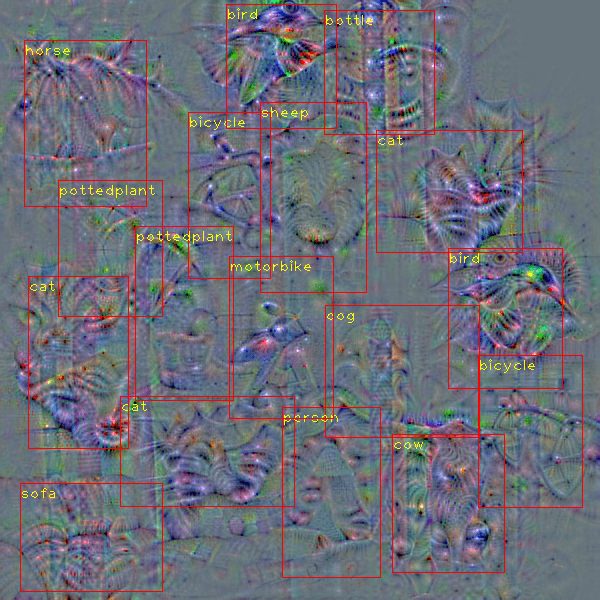}
    \end{minipage}     
    \\
    \pagebreak
    \vspace{0.2in}
    \textbf{\large{Pseudo-Targets}} & \textbf{\large{Background Initialization}}
\\
\\
         \begin{minipage}{.45\textwidth}
      \includegraphics[width=\linewidth]{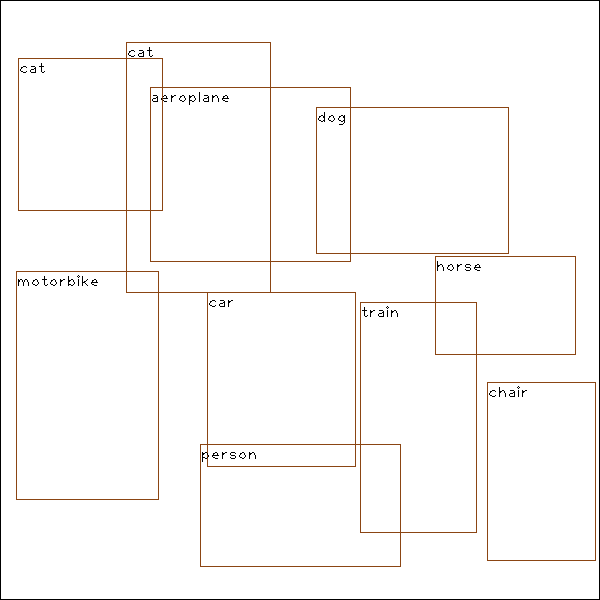}
    \end{minipage}  &               \begin{minipage}{.45\textwidth}
      \includegraphics[width=\linewidth]{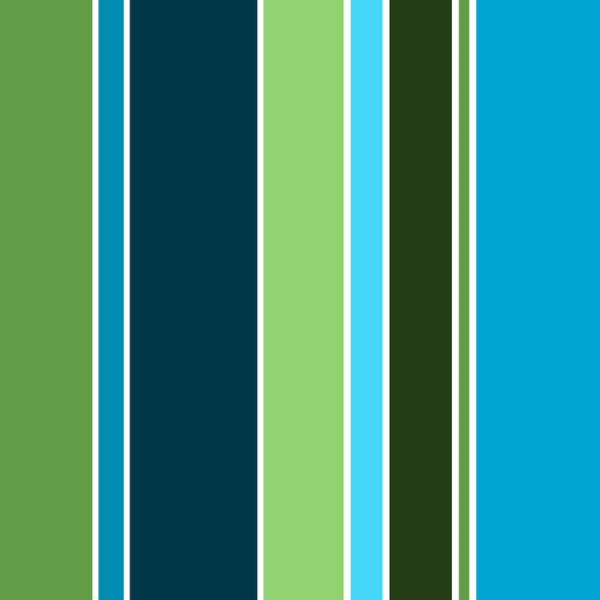}
    \end{minipage}                          \\
    \vspace{0.3in}
    & \\
    & \\
    \vspace{0.2in}
    \textbf{\large{MOI}} & \textbf{\large{MOI without background}}
    \\
    \\
                              \begin{minipage}{0.45\textwidth}
      \includegraphics[width=\linewidth]{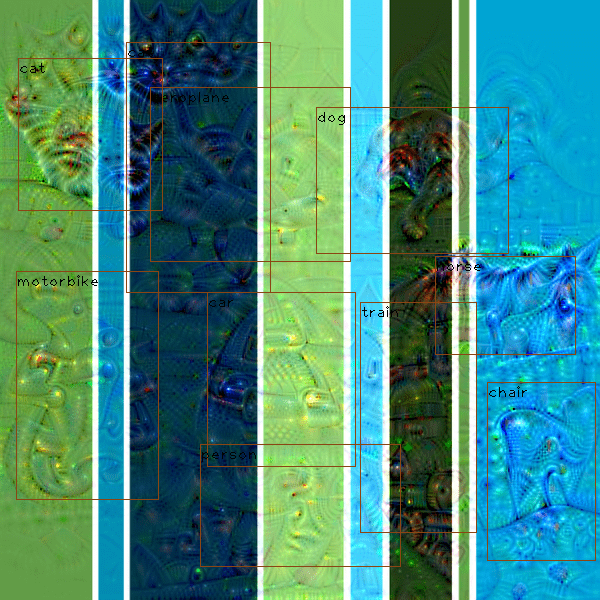}
    \end{minipage}                       &      \begin{minipage}{.45\textwidth}
      \includegraphics[width=\linewidth]{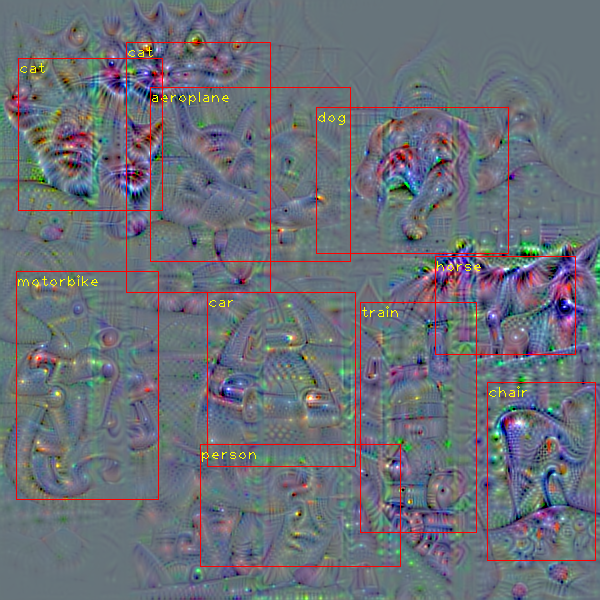}
    \end{minipage}     
    \\
    
    \pagebreak
    \vspace{0.2in}
    \textbf{\large{Pseudo-Targets}} & \textbf{\large{Background Initialization}}
\\
\\
         \begin{minipage}{.45\textwidth}
      \includegraphics[width=\linewidth]{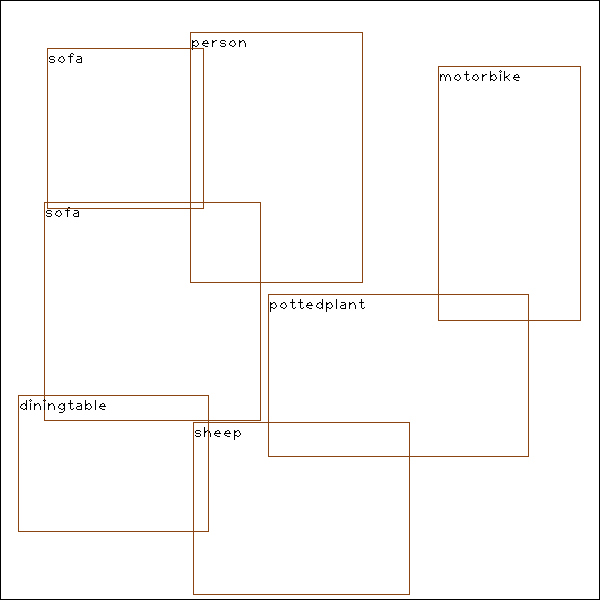}
    \end{minipage}  &               \begin{minipage}{.45\textwidth}
      \includegraphics[width=\linewidth]{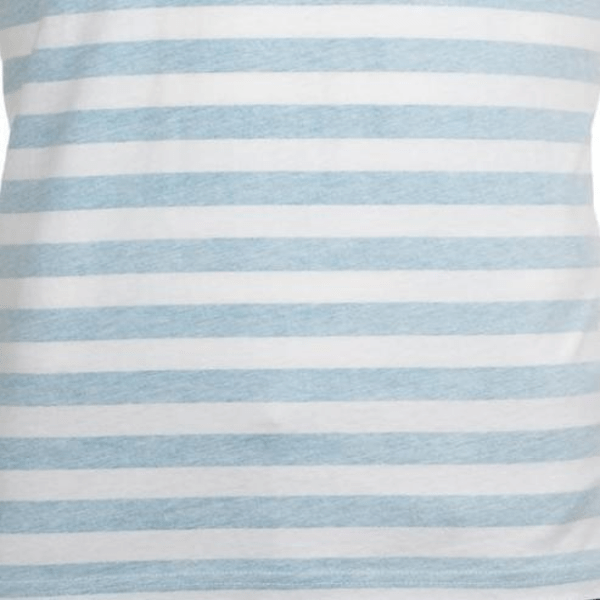}
    \end{minipage}                          \\
    \vspace{0.3in}
    & \\
    & \\
    \vspace{0.2in}
    \textbf{\large{MOI}} & \textbf{\large{MOI without background}}
    \\
    \\
                              \begin{minipage}{0.45\textwidth}
      \includegraphics[width=\linewidth]{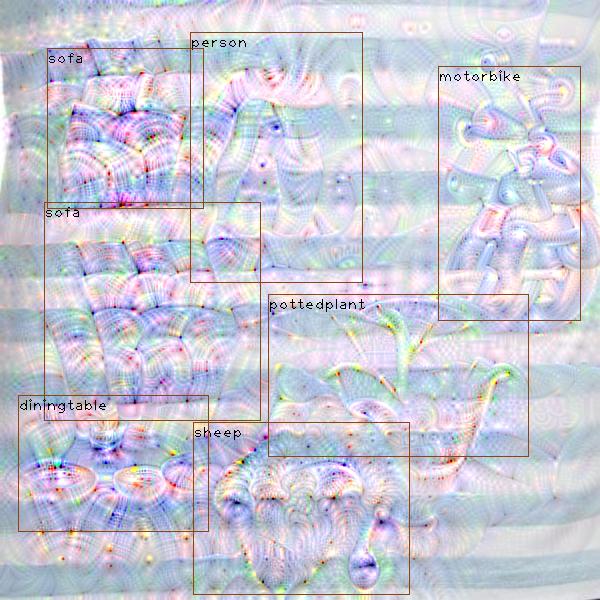}
    \end{minipage}                       &      \begin{minipage}{.45\textwidth}
      \includegraphics[width=\linewidth]{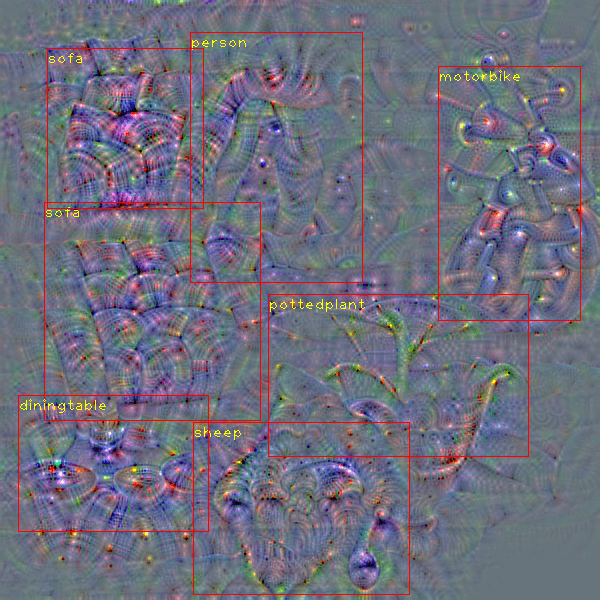}
    \end{minipage}     
    \\
    
    \pagebreak
    \vspace{0.2in}
    \textbf{\large{Pseudo-Targets}} & \textbf{\large{Background Initialization}}
\\
\\
         \begin{minipage}{.45\textwidth}
      \includegraphics[width=\linewidth]{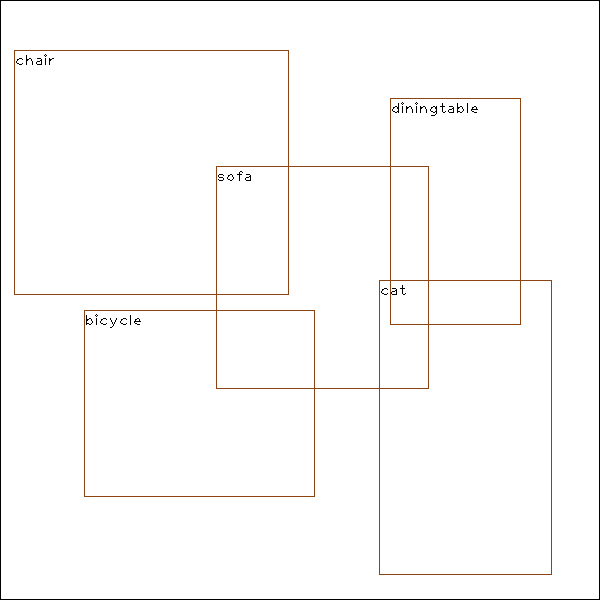}
    \end{minipage}  &               \begin{minipage}{.45\textwidth}
      \includegraphics[width=\linewidth]{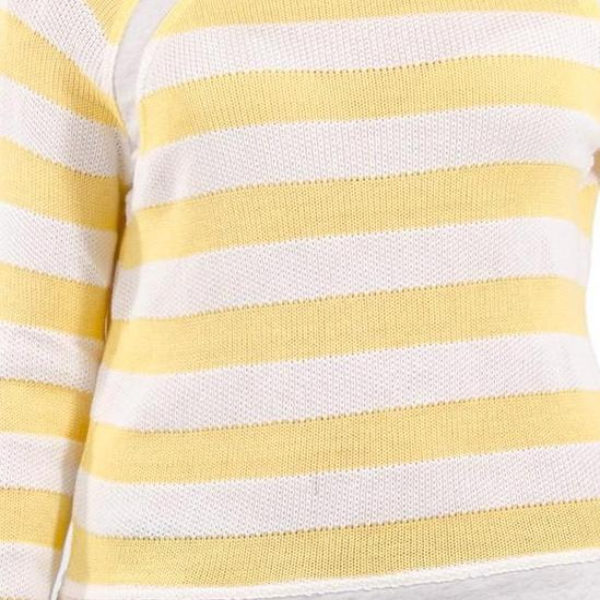}
    \end{minipage}                          \\
    \vspace{0.3in}
    & \\
    & \\
    \vspace{0.2in}
    \textbf{\large{MOI}} & \textbf{\large{MOI without background}}
    \\
    \\
                              \begin{minipage}{0.45\textwidth}
      \includegraphics[width=\linewidth]{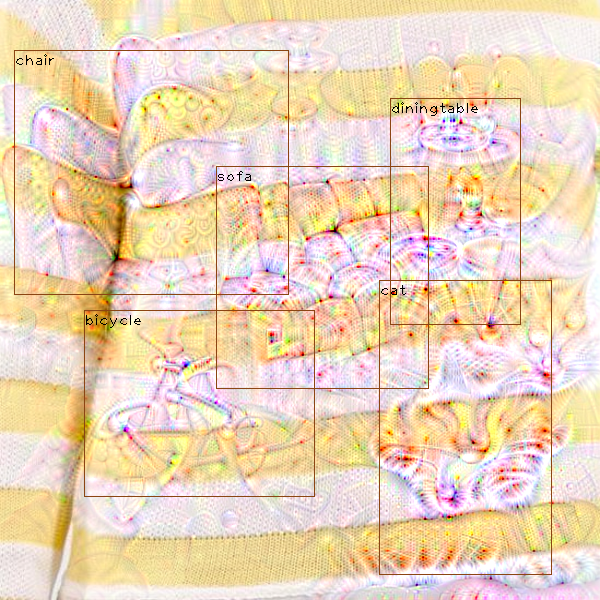}
    \end{minipage}                       &      \begin{minipage}{.45\textwidth}
      \includegraphics[width=\linewidth]{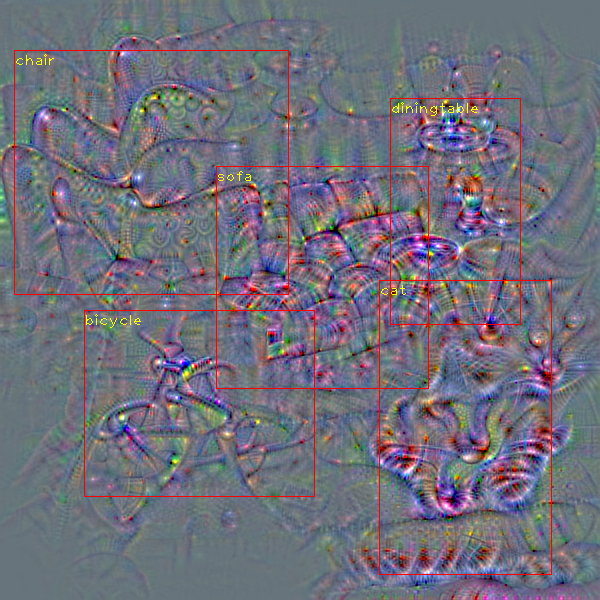}
    \end{minipage}     
    \\
    \pagebreak
    \vspace{0.2in}
    \textbf{\large{Pseudo-Targets}} & \textbf{\large{Background Initialization}}
\\
\\
         \begin{minipage}{.45\textwidth}
      \includegraphics[width=\linewidth]{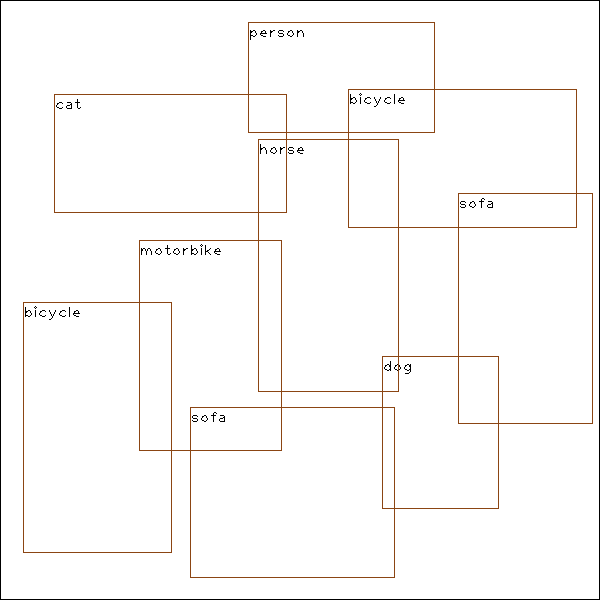}
    \end{minipage}  &               \begin{minipage}{.45\textwidth}
      \includegraphics[width=\linewidth]{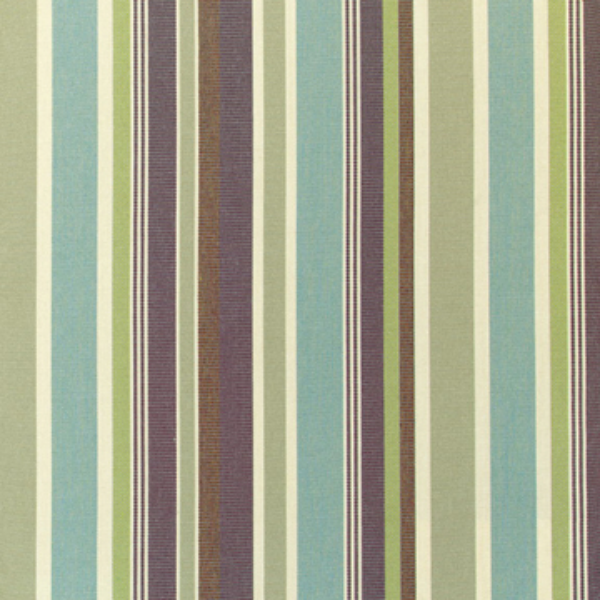}
    \end{minipage}                          \\
    \vspace{0.3in}
    & \\
    & \\
    \vspace{0.2in}
    \textbf{\large{MOI}} & \textbf{\large{MOI without background}}
    \\
    \\
                              \begin{minipage}{0.45\textwidth}
      \includegraphics[width=\linewidth]{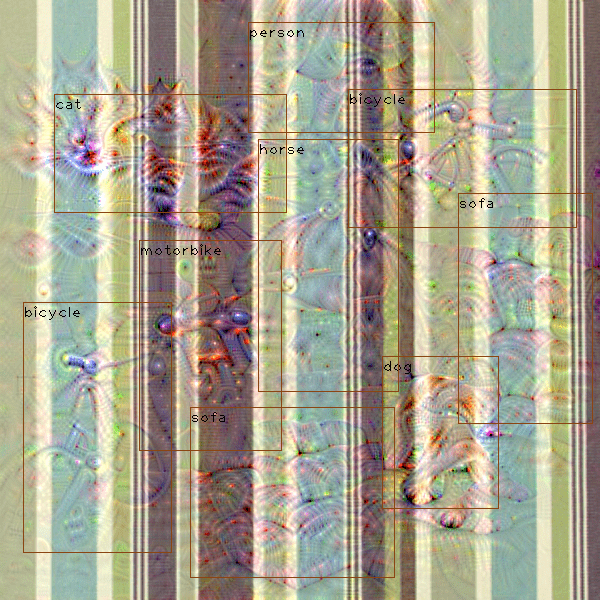}
    \end{minipage}                       &      \begin{minipage}{.45\textwidth}
      \includegraphics[width=\linewidth]{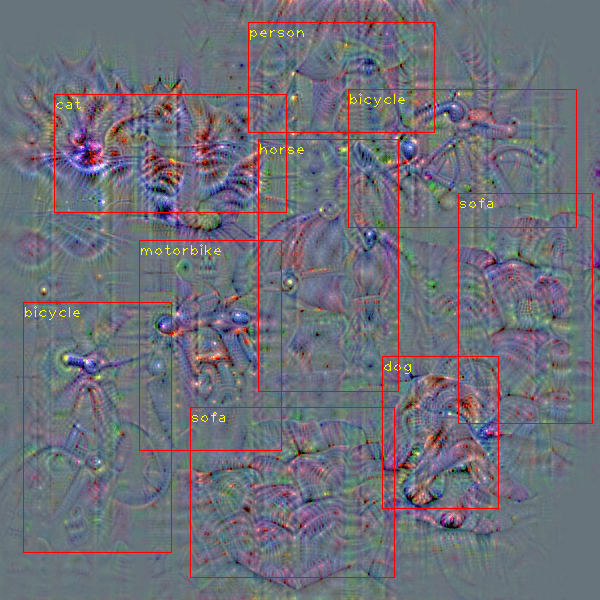}
    \end{minipage}     
    \\
    \pagebreak
    \vspace{0.2in}
    \textbf{\large{Pseudo-Targets}} & \textbf{\large{Background Initialization}}
\\
\\
         \begin{minipage}{.45\textwidth}
      \includegraphics[width=\linewidth]{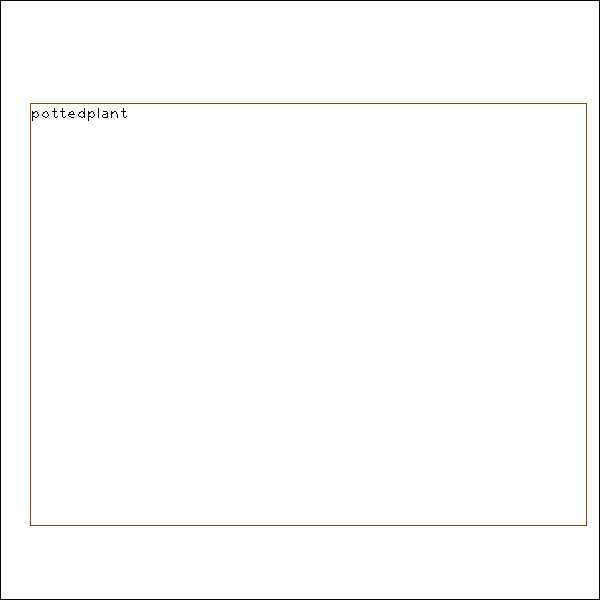}
    \end{minipage}  &               \begin{minipage}{.45\textwidth}
      \includegraphics[width=\linewidth]{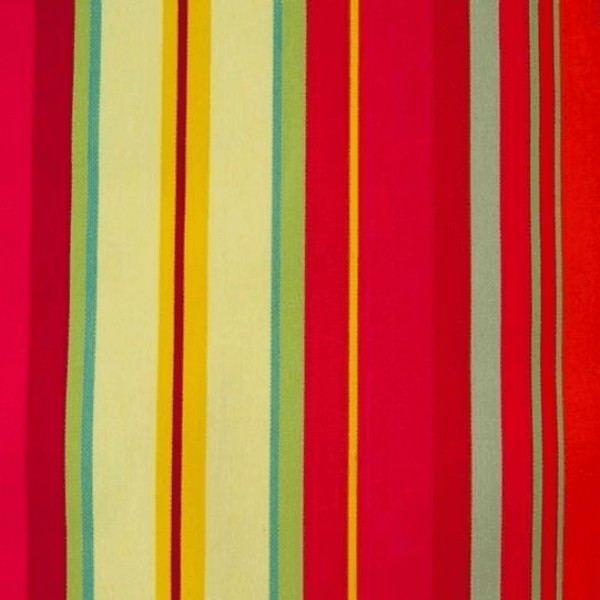}
    \end{minipage}                          \\
    \vspace{0.3in}
    & \\
    & \\
    \vspace{0.2in}
    \textbf{\large{MOI}} & \textbf{\large{MOI without background}}
    \\
    \\
                              \begin{minipage}{0.45\textwidth}
      \includegraphics[width=\linewidth]{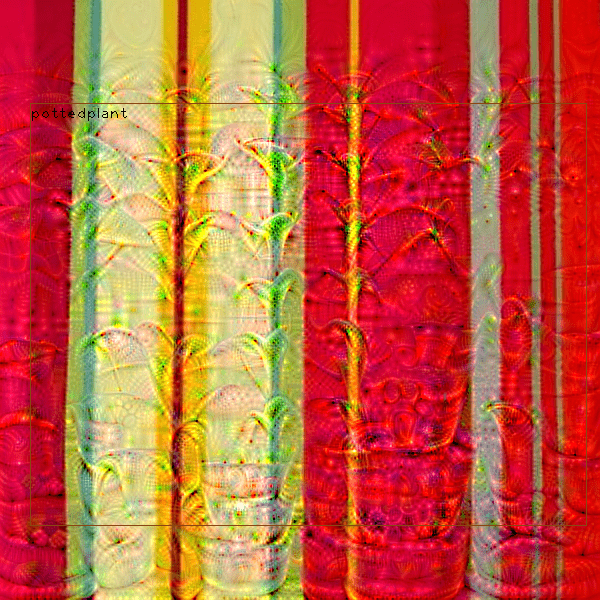}
    \end{minipage}                       &      \begin{minipage}{.45\textwidth}
      \includegraphics[width=\linewidth]{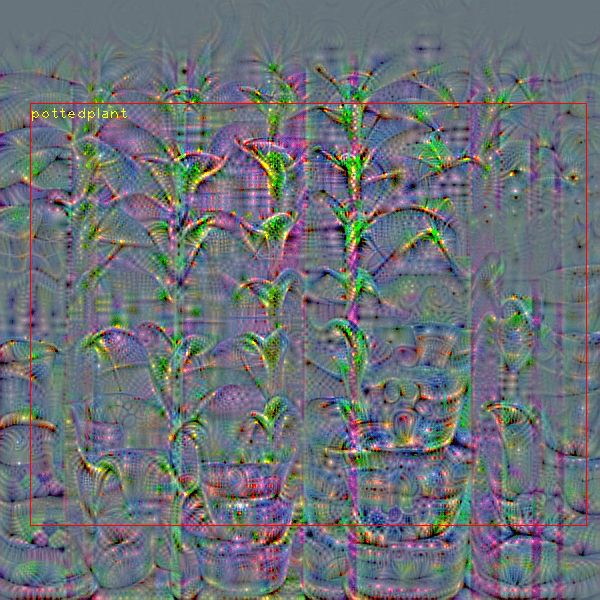}
    \end{minipage}     
    \\

\label{tab:my-table}
\end{longtable}

\newpage
\section{Hyperparameter Details}
\label{hyperparams}
The hyperparameters used in experiments can be divided into three categories : 
\begin{itemize}
\item Generation of MOIs 
\begin{table}[htp]
\centering
\scalebox{0.9}{
\begin{tabular}{|c|c|c|c|}
\hline
\multirow{2}{*}{Hyperparameters} & \multicolumn{3}{c|}{Generation of MOIs from the \Te{} network} \\ \cline{2-4} 
                                   & Resnet-18 (KITTI) & VGG-16 (Pascal) & Resnet-34 (Pascal) \\ \hline \hline
Number of distinct initializations & 5000              & 5000            & 5000               \\ \hline
Number of samples ($K$)              & 15000             & 15000           & 15000              \\ \hline
Augmentation                       & Flip, Cutout              & Flip, Cutout            & Flip, Cutout               \\ \hline
Maximum number of objects ($M$)      & 20                & 20              & 20                   \\ \hline
Imagenet Pretrained Model          & Pytorch           & Pytorch         & Pytorch                   \\ \hline
Learning Rate                      & 0.01              & 0.01            & 0.01                   \\ \hline
Optimizer                          & Adam              & Adam            & Adam                   \\ \hline
Batch size                         & 16                 & 4               & 8                    \\ \hline
Maximum Iterations                 & 100               & 100             & 100                   \\ \hline
Weight to diversity loss ($\lambda$)  & 2                 & 1.5               & 1.5                   \\ \hline
$IOU_{threshold}$                     & 0.1               & 0.1             & 0.1                   \\ \hline
\end{tabular}
}
\caption{Details of the hyperparameters used in the generation of MOIs.}
\label{tab:gen}
\end{table}
\item Distillation with MOIs 
\begin{table}[htp]

\scalebox{0.78}{
\centering
\begin{tabular}{|c|c|c|c|}
\hline
\multirow{2}{*}{Hyperparameters} &
  \multicolumn{3}{c|}{Distillation using MOIs as transfer set in the absence of original training data} \\ \cline{2-4} 
 &
  \begin{tabular}[c]{@{}c@{}}Resnet-18 to Resnet-18-half\\  (KITTI)\end{tabular} &
  \begin{tabular}[c]{@{}c@{}}VGG-16 to VGG-16 \\ (Pascal)\end{tabular} &
  \begin{tabular}[c]{@{}c@{}}Resnet-34 to Resnet-18 \\ (Pascal)\end{tabular} \\ \hline \hline
Number of samples                & 15000 & 15000 & 15000 \\ \hline
Total samples with augmentation  & 30000 & 30000 & 30000 \\ \hline
Weight to feature imitation loss & 5     & 0.01   & 1     \\ \hline
\end{tabular}
}
\caption{Details of hyperparameters used in distillation with MOIs as a transfer set.
}
\label{tab:distillation}
\end{table}
\item Teacher training with original data 
\begin{table}[htp]
\centering
\scalebox{0.9}{

\begin{tabular}{|c|c|c|c|}
\hline
\multirow{2}{*}{Hyperparameters} & \multicolumn{3}{c|}{\Te{} training on original data}    \\ \cline{2-4} 
                                 & Resnet-18 (KITTI) & VGG-16  (Pascal) & Resnet-34 (Pascal) \\ \hline \hline
Learning Rate (LR) & 0.001 & 0.001 & 0.001 \\ \hline
Step decay    & 5     & 6     & 6     \\ \hline
LR decay  & 0.1   & 0.1   & 0.1   \\ \hline
\end{tabular}
}
\caption{Hyperparameters involved in \Te{} training}
\label{tab:teacher_train}
\end{table}
\end{itemize}

\bibliography{references}
\end{document}